%% file: main.tex
\documentclass[11pt,a4paper]{article}

\usepackage{yemale}          %
\usepackage{etoc}            %

\newcommand{\paperfigure}[1]{\includegraphics{figures/#1.pdf}}

\hypersetup{
  pdftitle={Beyond Uncertainty Sets: Leveraging Optimal Transport to Extend Conformal Predictive Distributions to Multivariate Settings},
  pdfauthor={Eugene Ndiaye}
}

\title{Beyond Uncertainty Sets: Leveraging Optimal Transport to Extend Conformal Predictive Distributions to Multivariate Settings}

\author{%
  Eugene Ndiaye
  \thanks{
  Email: \email{ndiayeeugene@gmail.com}}
}

\date{}

\begin{document}

\maketitle

\input{subfiles/abstract}
\noindent The implementation is available at
\url{https://yemale.github.io/}.
\newpage
\etocsettagdepth{body}{subsubsection}
\etocsettagdepth{appendix}{none}
\tableofcontents
\newpage

\etocdepthtag.toc{body}
\input{subfiles/introduction}          %
\input{subfiles/background}            %
\input{subfiles/optimal_transport}     %
\input{subfiles/mcpd}                  %
\input{subfiles/geometry}              %
\input{subfiles/conformal_inference}   %
\bibliography{references}
\bibliographystyle{apalike}

\appendix
\etocdepthtag.toc{appendix}
\begingroup
  \etocsettagdepth{body}{none}
  \etocsettagdepth{appendix}{subsubsection}
  \renewcommand{\contentsname}{Appendix Contents}
  \tableofcontents
\endgroup
\newpage
\input{subfiles/proofs}
\input{subfiles/optimal_algorithm}

\end{document}

%% file: subfiles/abstract.tex
\begin{abstract}
Conformal prediction (CP) constructs uncertainty sets for model outputs with finite-sample coverage guarantees. 
Despite its major importance, it relies on ranking score functions that are only straightforward when the scores are scalar-valued which has limited CP to real-valued scores or ad-hoc reductions to one dimension. This limitation is critical, as vector-valued scores arise naturally in important settings such as multi-output regression or situations involving model aggregation, where each predictor in an ensemble provides its own score.
The problem of ordering vectors has been studied via optimal transport (OT), which provides a principled method for defining vector-ranks and (center-outward) multivariate quantile regions, though typically with only asymptotic coverage guarantees. Specifically, this loss of validity occurs because applying a single, fixed transport map, learned from finite data, to new test points introduces an uncontrolled approximation error. We restore finite-sample, distribution-free coverage by conformalizing the vector-valued OT quantile region. In our approach, a candidate's rank is defined via a transport map computed for the calibration scores augmented with that candidate's score—a step crucial for preserving validity. This defines a continuum of OT problems (one for each candidate) which appears computationally infeasible. However, we prove that the resulting optimal assignment is piecewise-constant across a fixed polyhedral partition of the score space. This allows us to characterize the entire prediction set tractably in $O(n^3)$ which matches the optimal complexity for solving a single assignment problem. Crucially, this also provides the machinery to address a deeper limitation of prediction sets: that they only indicate which outcomes are plausible, but not their relative likelihood.
In one dimension, conformal predictive distributions (CPDs) fill this gap by producing a predictive distribution with finite-sample calibration. Extending CPDs beyond one dimension remained an open problem. We construct, to our knowledge, the first multivariate CPDs with finite-sample calibration, i.e., they define a valid multivariate (center-outward) distribution where derived uncertainty region automatically has guaranteed (conformal) coverage. We present both conservative and exact randomized versions; the latter resulting in a multivariate generalization of the classical Dempster-Hill procedure. \\
\end{abstract}

%% file: subfiles/introduction.tex
\section{Introduction}
\label{sec:introduction}

When an AI model makes a prediction, can we attach a measure of uncertainty to that
prediction with a mathematically valid finite-sample guarantee? For example, a forecast of a stock’s future price or future temperature is never exact, because in part, the measurements on which it is based are themselves random. As a result, predicting a single value (the ``best guess'' given the data) provides no indication of the magnitude of the possible error margin. A range of plausible values, on the other hand, does provide such an indication, provided that we can determine the probability that it contains the true value.
Conformal prediction (CP)
\citep{gammerman1998learning,Vovk_Gammerman_Shafer05,Shafer_Vovk08}
provides a model-agnostic framework for quantifying uncertainty. It
transforms a point predictor into one that outputs a prediction set,
guaranteed to contain the true outcome with a user-specified probability
under exchangeability. Once coverage is guaranteed, the size of the set
indicates how confident the prediction is; see
\citet{correia2025informationtheoreticperspectiveconformal} for a detailed
information-theoretic perspective.
We consider a predictive model $\widehat y$ deployed on input--output pairs in
$\mathcal X\times\mathcal Y$ and a calibration set
\[
D_n=\{(X_1,Y_1),\ldots,(X_n,Y_n)\}.
\]
When the prediction model is deployed, we see a new input but not its corresponding outcome,
$$\bigl(X_{\text{new}},\cancel{Y_{\text{new}}}\bigr) = \bigl(X_{n+1},\cancel{Y_{n+1}}\bigr).$$
The whole construction is arranged around this missing entry and how uncertain we are about predicting that $\widehat y(X_{\text{new}}) \approx Y_{\text{new}}$.
We start with a user-defined non-conformity score
$$S(x,y;\widehat y)\in\mathbb R,$$ designed to measure the disagreement
between a model's prediction and an observed outcome. But since the target $Y_{\text{new}}$ is \emph{unknown}, we will replace it with one of its possible values $y$, calculate the error it would produce, and then compare this error to the calibration errors.
The uncertainty set $\Gamma_\alpha(x)$ is the set of values whose score is not atypically higher than the errors observed in the calibration data.
Concretely, the threshold for what is large or not is fixed as $Q_{n+1}(1-\alpha)$ the
$\lceil(n+1)(1-\alpha)\rceil$-th smallest calibration score and so
\begin{equation}
\label{eq:split_cp_set}
\Gamma_\alpha(X_{\text{new}})
=\big\{y\in\mathcal Y:S(X_{\text{new}},y;\widehat y)\leq Q_{n+1}(1-\alpha)\big\}.
\end{equation}
Choosing the magnitude of the residual as a score $S(x,y;\widehat y)=|y-\widehat y(x)|$ on
$\mathcal Y=\mathbb R$ gives the familiar interval
$
\Gamma_\alpha(X_{\text{new}})
=\bigl[\widehat y(X_{\text{new}}) \pm Q_{n+1}(1-\alpha)\bigr],
$
so the calibration quantile says how far to go on either side of the
prediction. The attractiveness of conformal prediction set is not only their simple construction but also they come with a strong coverage property.
If the calibration observations and the future observation are
exchangeable (their joint distribution is invariant under permutation) then the set in \Cref{eq:split_cp_set} contains the unobserved target $Y_{\text{new}}$ with probability $100(1-\alpha)\%$. And this holds for any distribution and any sample size $n$:
\begin{equation}
\label{eq:split_cp_coverage}
\mathbb P\big(Y_{\text{new}}\in\Gamma_\alpha(X_{\text{new}})\big)\geq1-\alpha.
\end{equation}

\paragraph{The Challenge of Vector-Valued Scores}
The core CP mechanism relies on ranking scores. While comparing real numbers is straightforward, there is no unique or
canonical way to order vectors. It poses a fundamental challenge of ``faithfully'' extending CP framework to vector-valued scores $$S(x, y; \widehat y) \in \mathbb R^d.$$\looseness=-1
\begin{center}
\emph{How can we generalize conformal prediction to handle vector-valued scores \\
while preserving its coverage guarantees?}
\end{center}

This question arises in several common contexts in machine learning. In multivariate regression, a
natural score is the full residual vector $S(x, y; \widehat y) = y - \widehat y(x) \in \mathbb{R}^d$, which captures the
direction and magnitude of the error across all output coordinates. For example, when the target is a location, our best guess may be incorrect for two values: latitude and longitude. 
Another important example is in classification, while a common scalar score measures the model's
confidence in the true class, it is a lossy summary that discards information about how the probability mass is distributed among the incorrect classes. Two predictions may both assign the same probability to
the true class, while one concentrates the remaining probability on a single
alternative and the other distributes it across multiple classes. 
To preserve this richer information about the distribution,
we can work directly with the full vector of predicted class probabilities. A natural vector score
can then be calculated from the difference between this probability vector and a “one-hot” vector
representing the true label. Furthermore, even when the response is univariate,
vector-valued scores arise in model aggregation where we want to combine the prediction of multiple models. Here, the individual prediction errors from an ensemble of
models can be concatenated into a single score vector, representing the collective judgment and
disagreement of the ensemble. This lets us
distinguish models that make similar errors in the same direction from
models whose errors point in different directions. Reducing these
vectors to a scalar, for instance by taking a norm, can discard semantics, 
directional or other structural information which we aim to retain in the conformal construction.

\paragraph{Beyond Prediction Sets}
Along with the question of how to construct a \emph{valid} uncertainty set, a second
question concerns the information it contains.
\begin{center}
\emph{How likely is a particular output within the conformal prediction set?}
\end{center}
A standard conformal prediction set does not directly answer this question. Among the plausible outcomes, the set itself provides no additional information that would allow us to distinguish their relative probabilities. To do so, we would instead need to define a
complete probability distribution over the space of outcomes, which provides a detailed likelihood for each outcome. This is typically useful to directly
guide decision-making. An uncertainty set can potentially support a decision based on the worst possible outcome it contains \citep{kiyani2025decision} while a predictive distribution
also allows probabilities to be assigned to events and consequences to
be averaged over possible outcomes \citep{hullman2025conformal}. It can therefore answer questions
that a prediction set alone leaves open. However, a minimal property we would request is that any prediction set extracted (by
selecting a region of a certain probability mass) automatically has valid conformal coverage as in \Cref{eq:split_cp_coverage}.

\paragraph{On Calibrated Predictive Distributions}

The classical Bayesian framework provides posterior predictive distribution
\citep{box1980sampling,rubin1984bayesianly,gelman1996posterior}.
It starts with a prior distribution over the model parameters and
updates it using the observed data. For each parameter value the model
gives a distribution for the future outcome, and these are averaged over
the posterior,
\begin{equation}
\label{eq:bayesian_predictive_distribution}
p(y\mid x_{n+1},D_n)
=
\int_\Theta p(y\mid x_{n+1},\theta)\,p(\theta\mid D_n)\,d\theta .
\end{equation}
This accounts for both the randomness of the future outcome and
uncertainty about the model parameters themselves.
However, prediction regions derived from this distribution do not, in
general, have finite-sample distribution-free coverage. Frequentist
guarantees are available for Bayesian procedures, but they are asymptotic
and rest on assumptions about the model and the prior. We refer to 
\citet{castillo2024bayesian} for a recent account of the frequentist
analysis of posterior distributions, uncertainty quantification included.
This disconnect has motivated work that
combines Bayesian prediction with conformal guarantees. For example,
\citet{fong2021conformal} use Bayesian predictive densities to construct
conformal prediction sets. This gives a way to use the model's
probabilities while retaining coverage, but it does not by itself
calibrate the full predictive distribution.
The question of what makes a predictive distribution \emph{``valid''} has a
long history. As emphasized by \citet{dawid1984present} and
\citet{diebold1997evaluating}, predictive distributions should be
evaluated together with the observations they predict. The probability
integral transform (PIT) provides a way to assess this calibration
\citep{hamill2001interpretation,gneiting2007probabilistic}. Building on
this foundation, \citet{shen2018prediction} formulate two requirements:
\begin{itemize}
\item For each new input, the procedure returns a proper cumulative
distribution function.
\item Its probability integral transform is uniformly distributed.
\end{itemize}
The second requirement connects the predictive distribution to
observations. For example, for a continuous predictive CDF, the future
outcome should fall below the reported $70\%$ quantile $70\%$ of the time
across repeated experiments. As we explain
in \Cref{sec:conformal_univariate}, this gives prediction intervals with
the coverage specified by their quantile levels.
Frequentist predictive distributions can also be constructed using
confidence distributions \citep{lawless2005frequentist,shen2018prediction}.
These have the same shape as
\Cref{eq:bayesian_predictive_distribution}, with the posterior replaced by
a confidence distribution $H(\theta;D_n)$ on the parameter space,
\[
G(y;x_{n+1},D_n)=\int_\Theta F_\theta(y\mid x_{n+1})\,dH(\theta;D_n).
\]
Their calibration depends on the model or on regularity conditions, and is
often justified asymptotically. Both integrate over $\Theta$, which is
rarely tractable when the parameter space is large. This leaves the
following question:
\begin{center}
\emph{Can we construct a predictive distribution with finite-sample calibration, \\
without assumptions on the data distribution beyond exchangeability?}
\end{center}

Just as conformal prediction extends a point prediction to a prediction
set, Conformal Predictive Distribution systems (CPDs) provide a way to report cumulative
values with a finite-sample calibration guarantee. This construction was
introduced by \citet{vovk2017nonparametric,vovk2018conformal,vovk2019universally}
for the univariate setting, and has since been developed in several
directions
\citep{vovk2020computationally,vovk2018cross,bostrom2021mondrian,johansson2023conformal}.
Conformal predictive distributions have also been compared with Bayesian predictive distributions \citep{lanngren2025conformal} and studied for Gaussian processes \citep{pion2025bayesian}.\looseness=-1

To our knowledge, the generalization of CPDs to multivariate settings was left open.\looseness=-1

\subsection{Where Optimal Transport Enters Conformal Prediction}

To see why OT is already present in CP, let us first return to the basic construction.
A distribution on the real line, represented by its CDF, tells us how much
probability lies below each value. Thus, it ranks values within the
population and places them on a common uniform scale from zero to one.
An uncertainty set collects plausible values for a future observation. We can construct one by reading this correspondence backward: choose a part of the
probability scale with the desired mass, then retain the values that the CDF
maps there. When the CDF is unknown but observations are available, probabilities are
replaced by frequencies. by arranging the $n$ observations from smallest to
largest, they divide the real line into $n+1$ intervals that describes successive quantile
positions.
Now, if the observations appear in no privileged order, then reordering the data does not change their joint distribution, which is exactly what exchangeability means. The CP insight is that the candidate \emph{must} therefore be included in the ranking itself to preserve this symmetry. When the candidate is the actual future observation, all $n+1$ observations participate in the same construction and are treated in the same way. Every possible rank is therefore equally likely, so each source cell contains the future observation with marginal probability $1/(n+1)$. The one-sided conformal set introduced in \Cref{eq:split_cp_set} is obtained by assembling these source cells from left to right until their combined probability reaches the desired coverage.

Optimal transport extends these operations to vectors. Among all applications that
map one distribution onto another, it selects the one that minimizes the
average distance between the source and target values.
In one dimension, sorting in ascending order is a solution to this optimization problem and for a continuous (resp. discrete) uniform target, this recovers the CDF (resp.normalized ranks i.e. empirical CDF evaluated at the observations).
The same minimization can be performed for vectors, even though there is
no longer left-to-right ordering. It extends the role that sorting operator played on the real line, leading to the notion of multivariate distribution and quantile
functions introduced in\citep{chernozhukov2017,hallin2021}.
Once again, prediction regions (aka quantile regions) follow by reading the map backward.
Choose a region containing $90\%$ of the probability mass according to the target distribution, and collect
the source values that are transported there. By the definition of a transport map, these values should 
therefore form a prediction region with the same coverage, 
i.e., $90\%$ of the probability mass of the original distribution. Nevertheless, when the matching uses only the calibration observations, it
transports their empirical distribution exactly while leaving the future
observation outside the construction. This breaks the symmetry  and the uncertainty set obtained no longer give coverage for the future observation!

\subsection{Contributions Overview}

We revisit the conformal inference framework through the lens of optimal
transport. We introduce candidate-augmented optimal transport: each
candidate participates in the matching alongside the calibration
observations. At the true future outcome, all observations participate
symmetrically and every reference label is used once. Exchangeability
therefore makes the future observation equally likely to receive any
label. The accepted reference labels determine finite-sample coverage.
This provides a common construction for vector-valued ranks, prediction sets, and predictive distributions.
We subsume our contributions as follows:
\begin{itemize}
\item \emph{Conformal Prediction with Vector-Valued Scores.}
We use optimal transport to assign vector-valued scores to positions in a
reference space. These positions serve as multivariate rankings, and the prediction regions
are naturally formed by selecting the highest-ranked vectors, just as scalar rankings do in classical conformal prediction. The key difference here is that the cumulative distribution is evaluated directly on the
full-score vectors, without first reducing them to a scalar summary. In that sense, it extends the notion of "conformalization" itself to vectors.

\item \emph{A Tractable Algorithm.}
The augmented-transport defines one transport problem per candidate, hence infinitely many; which is computationally infeasible at first sight. We solve this by proving that
the optimal assignment function is piecewise-constant across a fixed polyhedral partition of
the source space.
More precisely, we describe the whole family through $n+1$ fixed costs. For each reference
label, its cost is the smallest cost of assigning the calibration scores to
the remaining labels.  We show how to obtain all of them from
a \emph{single} assignment solve followed by one shortest-path pass. Hence a total cost of $O(n^3)$.\looseness=-1

\item \emph{Multivariate Conformal Predictive Distributions.}
We extend the scalar conformal predictive construction of
\citet{vovk2017nonparametric} to vectors. We show that the newly introduced (augmented) assignments
obtained for different candidates together define proper multivariate distributions, ranks and quantiles à la \citet{hallin2021}. By that we mean, it defines a monotone map,
extending the increasing CDF on the line to the multivariate setting. At the future observation, this map has exactly
the discrete reference distribution. To obtain the continuous reference
distribution, we divide the reference space into cells of equal probability
and draw within the assigned cell according to the reference law.
This gives a multivariate randomized PIT with exact finite-sample
calibration. Now, it suffices to choose 
how to distribute probability inside each cell to complete a predictive
distribution from which we can sample. This formally extends the Dempster--Hill
construction \citep{dempster1963direct,hill1968posterior} to vectors.
\looseness=-1
\end{itemize}

\subsection{Related Work}
Work on multivariate uncertainty for conformal prediction has followed several practical paths. We briefly review the main lines; further discussion and implementation details appear in the appendix. Our approach avoids learning a global transport map, includes the test candidate within the OT formulation to retain exact finite-sample guarantees, removes the need to solve OT for each candidate by exploiting a polyhedral partition of score space, and provides, to our knowledge, the first multivariate CPDs with finite-sample calibration.

\paragraph{Scalarizing vector scores with norms.}
A straightforward approach reduces a vector of residuals to a single number using a norm, so that standard univariate conformal prediction can be applied ``as is.'' Variants first re-scale or whiten the coordinates to account for different scales and correlations across targets, for example by using a Mahalanobis-type norm \citep{pmlr-v152-johnstone21a,messoudi2022ellipsoidal}. More recent work learns an ellipsoidal shape to shrink the prediction sets while preserving coverage \citep{braun2025minimum}. A broad benchmark is provided by \citet{dheur2025multioutputconformalregressionunified}. These methods are simple and valid, but they can discard directional information and impose rigid geometry on the resulting sets.
Among them, the
\emph{Optimal-transport merging with a pre-fit transform} initially appears as a pragmatic alternative learns a fixed transport map on a hold-out split that sends multivariate scores to a simple reference distribution, and then calibrates a scalar summary of the transformed scores with standard conformal prediction \citep{thurin2025optimaltransportbasedconformalprediction,klein2025multivariate}. To maintain exchangeability, data are typically split three ways: one part to train the predictor, one part to fit the transport, and one part to calibrate. This preserves finite-sample validity because the transport is fixed before calibration. However, it is not equivalent to exact conformal prediction, even in one dimension, since the test point is not included when defining the transform.

\paragraph{Vector quantile regression.}
Vector quantile regression and related vector-quantile methods estimate conditional quantile maps using optimal transport or geometric multivariate quantiles \citep{chaudhuri1996geometric,serfling2002quantile,carlier2016vector,rosenberg2022fast,pegoraro2023vector}. These methods target conditional distributional modeling, whereas our construction starts from arbitrary vector-valued nonconformity scores and builds finite-sample conformal ranks. When finite-sample coverage is desired, VQR-type regions are typically followed by an additional scalar conformalization step \citep{feldman2023calibrated}.

\paragraph{Flow-based conformal predictive distributions.}
Recent work has proposed flow-based conformal predictive distributions for high-dimensional and structured output spaces \citep{harris2026flow}. Given a differentiable nonconformity score and conformal thresholds $\tau_\alpha$, this approach constructs a deterministic flow whose trajectories converge to conformal boundaries
$
\partial C_\alpha(x)=\{y:S(x,y)=\tau_\alpha\}.
$
By mixing boundary measures over confidence levels $\alpha$, it yields a predictive measure whose empirical conformal sets are quantile regions of the constructed distribution. This is complementary to our goal. The flow-based construction is compatible with a conformal filtration of sets; it does not construct a candidate-wise PIT map whose evaluation at the true response is uniform. In contrast, our multivariate CPDs aim to generalize the randomized empirical PIT itself, replacing one-dimensional gaps by equal-mass reference cells.\looseness=-1

\paragraph{Copula-based multivariate modeling.}
Another line models the joint distribution of multivariate scores using copulas, after first mapping each coordinate to a uniform scale via empirical marginal distributions \citep{messoudi2021copula,park2024semiparametric}. This can produce geometrically adapted multivariate sets. The trade-off is validity: coverage becomes approximate and depends on how well the marginals and copula are estimated. Additional data splitting can be used to recover coverage guarantees \citep{sun2022copula}. Related copula approaches that preserve coverage have also been proposed \citep{mukama2024copula,mukama2025copula}. Similarly, max-rank methods aggregate per-dimension ranks and use a single, more conservative cutoff \citep{timans2025max}. These procedures deliver finite-sample guarantees for joint coordinate-wise errors, but they can be conservative and do not produce predictive distributions.\looseness=-1

%% file: subfiles/background.tex
\section{From Prediction Sets to Predictive Distributions: the Univariate Case}
\label{sec:conformal_univariate}
We revisit how conformal uncertainty sets arise from first principles.

\subsection{Uncertainty Sets from the Ground-Truth Distribution}
\label{subsec:from_ground_truth_to_empirical_prediction_sets}

Suppose the distribution $\mu$ of the random variable $Z$ we want to predict
is \emph{known}, and we seek an interval containing its value with a prescribed
probability.
We leverage the cumulative distribution function
$F^\star(t)=\mu((-\infty,t])=\mathbb P\{Z\leq t\}$, the proportion of the
population at or below $t$. To highlight the principle, assume for
simplicity that $F^\star$ is continuous and strictly increasing. The events
$\{Z\leq t\}$ and $\{F^\star(Z)\leq F^\star(t)\}$ are then the same, so
\[
  \mathbb P\{F^\star(Z)\leq F^\star(t)\}=\mathbb P\{Z\leq t\}=F^\star(t).
\]
Continuity makes every $u\in(0,1)$ of the form $F^\star(t)$, so this reads
$\mathbb P\{F^\star(Z)\leq u\}=u$ for every such $u$: the transformed observation
\begin{equation}
\label{eq:true_pit}
U:=F^\star(Z)\sim\mathrm{Uniform}[0,1]
\end{equation}
has a distribution we know, whatever $\mu$ is. This is the probability
integral transform.
Here, it is crucial to note that the transformation depends on $\mu$, but the distribution it produces does not depend on it: a single
rule applies to all distributions. Consequently, no assumptions are made regarding the form of the
given distributions—in particular, whether they are Gaussian or have any other parametric form
that might not hold up in practice. This property is called “\emph{distribution-free}.”
We set out to measure with $\mu$, the object we do not know:
$F^\star(t)=\mu((-\infty,t])$. Since $U$ is uniform, after the transform we
measure instead with the uniform measure $\nu$ on $[0,1]$, which assigns to
an interval its length.
The two measurements agree through $F^\star$: for $[a,b]\subseteq[0,1]$,
\[
  \mu\bigl((F^\star)^{-1}([a,b])\bigr)
  =\mathbb P\{F^\star(Z)\in[a,b]\}
  =\nu([a,b])
  =b-a.
\]
So to get coverage $1-\alpha$, we pick an interval of length $1-\alpha$.
The central choice
$B_\alpha:=[\alpha/2,1-\alpha/2]$ returns the familiar quantile interval
$[q_{\alpha/2},q_{1-\alpha/2}]$, where $q_u:=(F^\star)^{-1}(u)$; see
\Cref{fig:scalar_cdf_construction}(a).
More generally, we can keep the values $z$ whose transform lands in a
reference set $B$. This is the \emph{pullback} of $B$, and
\Cref{eq:true_pit} gives its coverage:
\begin{align}
\mathcal Q_{F^\star}(B)&=\{z\in\mathbb R:F^\star(z)\in B\}=(F^\star)^{-1}(B),
\label{eq:true_confidence_set}\\
\mathbb P\{Z\in\mathcal Q_{F^\star}(B)\}&=\mathbb P\{F^\star(Z)\in B\}=\nu(B).
\label{eq:true_pullback_coverage}
\end{align}
So any $B$ with $\nu(B)=1-\alpha$ gives a region of coverage $1-\alpha$.
\subsection{The Dempster--Hill Construction}
\label{subsec:dempster_hill}

In practice, $F^\star$ is unknown. We observe $Z_1,\ldots,Z_n$, indexed by
$[n]:=\{1,\ldots,n\}$, and want to predict the next observation
$Z_{n+1}$, all with the same law $\mu$. 
Since a point estimate (the best estimate based on the observations) will not be
exact, it is natural to seek a range of plausible values. Suppose that $F^\star$ is continuous so that the probability that two observations coincide is
zero. We will therefore assume that $Z_1,\ldots,Z_{n+1}$ are all pairwise distinct
, and refer to their ranks without further specification.
Without $F^\star$ we estimate probabilities by counting: the empirical
distribution places a unit mass $\delta_{Z_i}$ at each observation, with
weight $1/n$:
\begin{equation}
\label{eq:empirical_cdf}
\mu_n=\frac1n\sum_{i=1}^n\delta_{Z_i},
\qquad
F_n(t)=\mu_n((-\infty,t])
=\frac{\#\{i\in[n]:Z_i\leq t\}}{n}.
\end{equation}
Thus $F_n(t)$ is the fraction of observations at or below $t$. Replacing
$F^\star$ by $F_n$ gives the region
\begin{equation}
\label{eq:plugin_empirical_region}
\mathcal Q_{F_n}(B)=\{z\in\mathbb R:F_n(z)\in B\}=F_n^{-1}(B).
\end{equation}
However, this replacement does not preserve the PIT relation. Simply because
$F_n(Z_{n+1})$ takes values in the finite grid $\{0,1/n,\ldots,1\}$,
so it cannot follow the continuous uniform law $\nu$. The length (measured with $\nu$) of $B$
therefore no longer directly determines coverage. \\

\paragraph{Augmented Samples}
Conformal prediction
\citep{gammerman1998learning,Vovk_Gammerman_Shafer05,Shafer_Vovk08}
obtains an exact reference law by including the candidate in the counting
operation itself.
For a candidate value $z\in\mathbb R$, form the augmented vector
\begin{equation}
\label{eq:candidate_augmented_score_vector}
\zeta(z):=(Z_1,\ldots,Z_n,z),
\qquad
\zeta_i(z)=
\begin{cases}
Z_i, & i\in[n],\\
z,   & i=n+1.
\end{cases}
\end{equation}
Its empirical distribution gives each of its $n+1$ components weight
$1/(n+1)$:
\begin{equation}
\label{eq:augmented_empirical_cdf}
\mu_{\zeta(z)}=\frac1{n+1}\sum_{i=1}^{n+1}\delta_{\zeta_i(z)},
\qquad
F_{\zeta(z)}(t)=\mu_{\zeta(z)}((-\infty,t]).
\end{equation}
Evaluating this CDF at the candidate itself gives
\begin{equation}
\label{eq:augmented_empirical_score_transform}
\begin{aligned}
F_{n+1}(z):=F_{\zeta(z)}(z)
&=\frac{\#\{i\in[n+1]:\zeta_i(z)\leq\zeta_{n+1}(z)\}}{n+1}
\end{aligned}
\end{equation}

\begin{figure}[tb]
\centering
\paperfigure{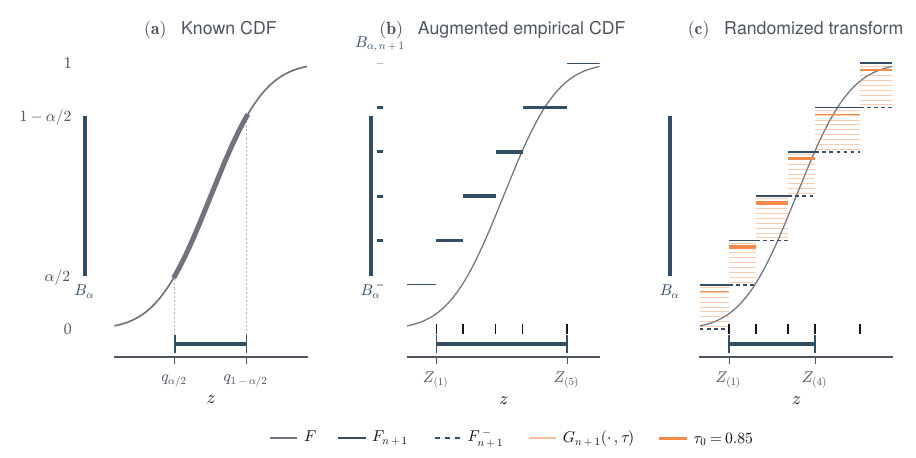}
\caption{\textbf{Choosing and pulling back the reference set.}
Left: for a known CDF, $B_\alpha$ is chosen by length and pulls back to a
quantile interval. Center: under the discrete law $\nu_{n+1}$,
$B_{\alpha,n+1}$ is chosen by counting labels; here five observations are
drawn from $N(0,9)$ and
$\alpha=0.4$, so four of the six labels are retained. Right: randomization
fills the reference intervals and restores $\nu$; the bold orange curve
shows the realization $\tau_0=0.85$. The horizontal brackets are the
corresponding pullbacks on the $z$-scale.}
\label{fig:scalar_cdf_construction}
\end{figure}

At $z=Z_{n+1}$, the
augmented vector becomes the complete sample.
The numerator is the rank $R_{n+1}$ of the future observation ie number of augmented observations at or below it. The augmented transform is therefore the rank divided by the sample size:
\[
F_{n+1}(Z_{n+1})=\frac{R_{n+1}}{n+1}.
\]
If the observations are exchangeable, every rank is equally likely, so
\[
U_{n+1}:=F_{n+1}(Z_{n+1})
\sim
\mathrm{Uniform}\left\{\tfrac1{n+1},\tfrac2{n+1},\ldots,1\right\}
=:\nu_{n+1}.
\]
This is the empirical counterpart of \Cref{eq:true_pit}. The transformed
future observation again has a distribution we know, whatever $\mu$ is;
it is now carried by $n+1$ equally likely reference points rather than by
the whole interval.
Write $u_j:=j/(n+1)$ for these points, so that
$F_{n+1}(Z_{n+1})=u_{R_{n+1}}$. The transform does more than produce a law:
it matches the $n+1$ augmented observations to the $n+1$ reference points,
sending the observation of rank $j$ to $u_j$. Ordering the
calibration observations as
\begin{equation}
\label{eq:scalar_order_scores}
Z_{(1)}<\cdots<Z_{(n)},
\qquad Z_{(0)}:=-\infty,\quad Z_{(n+1)}:=+\infty,
\end{equation}
gives the empirical quantiles, and they cut the line into $n+1$ source
gaps
\begin{equation}
\label{eq:scalar_order_gaps}
V_j:=\bigl(Z_{(j-1)},Z_{(j)}\bigr),
\qquad j\in[n+1],
\end{equation}
running from $V_1=(-\infty,Z_{(1)})$ to $V_{n+1}=(Z_{(n)},+\infty)$. 
Choose a fixed
reference set $B\subseteq[0,1]$, and retain the candidates
whose transformed values belong to it:
$$\mathcal Q_{F_{n+1}}(B)=\{z:F_{n+1}(z)\in B\}.$$ Its coverage is
\begin{equation}
\label{eq:univariate_conformal_score_coverage}
\mathbb P\{Z_{n+1}\in\mathcal Q_{F_{n+1}}(B)\}
=\nu_{n+1}(B)
=\frac{\#\{j\in[n+1]:j/(n+1)\in B\}}{n+1}.
\end{equation}
Under $\nu$ the mass of a set was its length; under $\nu_{n+1}$ it is the
fraction of reference labels the set contains. 
The appropriate measure here is therefore the
discrete length, and the procedure remains the same: choose a set with a mass at least equal to
$1-\alpha$ on the reference scale, then shift it backward.
Each rank index has a mass equal to $1/(n+1)$, so reaching
$1-\alpha$ means retaining cumulatively at least $(n+1)(1-\alpha)$. For example, by symmetrically eliminating the tails at both ends, we obtain the central set $B_{\alpha,n+1}$, as shown in
\Cref{fig:scalar_cdf_construction}(b).

Something has been lost along the way. There are only $n+1$ reference
points, so only a finite level of reachable coverages $u_1,\ldots,u_{n+1}$ can be attained. A
target $1-\alpha$ falling strictly between two of them cannot be met: we
have to climb to the next point up, and the region we return covers more
than was asked. The overshoot is a failure of the
scale, which offers no value between $u_{j-1}$ and $u_j$. A fix is to spread that mass over the interval it ends. Let's denote
\begin{equation}
  \label{eq:scalar_reference_intervals}
I_j=[u_{j-1},u_j],
\qquad j\in[n+1],
\quad u_0:=0,
\end{equation}
one interval of length $1/(n+1)$ per reference point, together partitioning
$[0,1]$ up to endpoints. Rank $j$ now selects the whole segment $I_j$, and rather than
returning a single endpoint $u_j$, we can draw uniformly inside; see \Cref{fig:scalar_cdf_construction}(c) for an illustration. The draw is
\begin{equation}
\label{eq:randomized_empirical_pit}
G_{n+1}(z,\tau)
=(1-\tau)\,u_{j-1}+\tau\,u_j =(1-\tau)F_{n+1}^-(z)+\tau F_{n+1}(z).
\end{equation}
where $\tau\sim\operatorname{Uniform}(0,1)$ and
\[
F_{n+1}^-(z):=F_{\zeta(z)}(z-)
=\frac{\#\{i\in[n+1]:\zeta_i(z)<\zeta_{n+1}(z)\}}{n+1}.
\]
This is the randomized PIT used in scalar (smoothed) conformal predictive
systems
\citep{vovk2017nonparametric,vovk2018conformal,vovk2019universally}.
As $\tau$ runs over $[0,1]$ the value moves continuously from $u_{j-1}$ to
$u_j$, so the pair $(z,\tau)$ can reach every level.
We summarize these already known results as follow.

\begin{proposition}[Discrete and randomized augmented PIT]
\label[proposition]{lm:PIT_onedim}
\label[proposition]{lm:randomized_PIT}
Let $Z_1,\ldots,Z_{n+1}$ be exchangeable real-valued random variables with
no ties, and let $\tau\sim\operatorname{Uniform}(0,1)$ be independent of
them. Then the augmented transform is calibrated on the discrete reference
law, and its randomization on the continuous one:\looseness=-1
\begin{align}
F_{n+1}(Z_{n+1})&\sim\nu_{n+1},
\label{eq:discrete_empirical_pit}
\\
G_{n+1}(Z_{n+1},\tau)&\sim
\frac{1}{n+1}\sum_{j=1}^{n+1}\nu(\cdot\mid I_j)=\nu,
\label{eq:univariate_uniform_mixture}
\end{align}
where $\nu(\cdot\mid I_j) = \operatorname{Uniform}(I_j)$.
\end{proposition}

\begin{remark}
Conformal prediction is in this sense an alternative to controlling the
whole process $\sup_{t\in\mathbb R}|F_n(t)-F^\star(t)|$: instead of making $F_n$
close to $F^\star$ everywhere and paying for the residual error, it changes the
reference law so that the transformed observation has a known law exactly;
\cite{JohnstoneNdiaye2026} develops the comparison.
\end{remark}

The reference partition $I_1,\ldots,I_{n+1}$ in \Cref{eq:scalar_reference_intervals}matches the source gaps
$V_1,\ldots,V_{n+1}$ of \Cref{eq:scalar_order_gaps} one for one. Order
statistics have long been used to construct distribution-free tolerance
intervals \citep{wilks1941determination}. A candidate in
$V_{i+1}=(Z_{(i)},Z_{(i+1)})$ has exactly $i$ calibration observations
below it, so
$
G_{n+1}(z,\tau)=\frac{i+\tau}{n+1}.
$
Randomization therefore carries $V_j$ onto the whole of $I_j$. From now on, the transform maps a candidate $z$ to 
$
\label{eq:scalar_predictive_band}
\mathcal I_{n+1}(z):=\bigl[F_{n+1}^-(z),F_{n+1}(z)\bigr]=I_{j(z)},
$
where $j(z)$ indexes the source interval (among the $V_j$) containing $z$, and the
randomization distributes $\nu(\,\cdot\mid I_{j(z)})$ over it.

\paragraph{Hill hypothesis: \emph{the next observation is equally likely to fall within any interval
between two consecutive order statistics}}
The precursor to conformal prediction is arguably Wilks’s methods \citep{wilks1941determination}, which subsequently gave rise to Hill’s so-called $A_{(n)}$ hypothesis \citep[see][]{dempster1963direct,hill1968posterior}. They intended to construct a predictive distribution without relying on the classical Bayesian approach, which uses a prior that could be considered "subjective". Assuming Hill's hypothesis and taking a central block of gaps
and reading it back on the original scale returns the classical tolerance
interval: the gaps $V_{k+1},\ldots,V_{n+1-k}$ carry
$n+1-2k$ units of mass, so
\[
\mathbb P\bigl\{Z_{n+1}\in[Z_{(k)},Z_{(n+1-k)}]\bigr\}
=\frac{n+1-2k}{n+1}.
\]
Thus taking $k=\lfloor\alpha(n+1)/2\rfloor$ gives coverage at least $1-\alpha$, provided $n+1\geq2/\alpha$; below that no interval of order statistics reaches the target. One can now easily see that this corresponds to uncertainty sets obtained from the randomized distribution. This was initially shown by \citet{vovk2017nonparametric}.
Note that these weights say how much mass each source gap receives; they do not say
how it is spread inside. Choosing for each $j$ a probability law
$\lambda_j$ carried by $V_j$ completes a law on the original line,\looseness=-1
\[
\Pi=\frac{1}{n+1}\sum_{j=1}^{n+1}\lambda_j ,
\]
and \emph{different choices of the $\lambda_j$ give different completions}.
In the target space, the filling scale is naturally specified; rank $j$ yields
$\nu(\,\cdot\mid I_j)=\operatorname{Uniform}(I_j)$, the conditional reference distribution
on the selected interval, and it is this choice that makes
the PIT exact. However, on the original scale, filling with $\lambda_j$ is arbitrary (or "subjective" as some would say...). In any case, its role is to shape the complete distribution $\Pi$, and does not alter the calibration.

\subsection{Conformal Prediction}
\label{subsubsec:conformal_predictive_distributions}

Nothing in the previous constructions used any property of $Z$ beyond the
exchangeability of the $n+1$ values. For supervised prediction we take that
function to be a score comparing an outcome with a model's prediction,
\[
Z_i=S(X_i,Y_i;\widehat y),
\qquad
S_x(y):=S(x,y;\widehat y),
\]
so that a candidate outcome $y$ at input $x$ enters through $S_x(y)$. This allows a large flexibility because we can virtually operates with any input/output and black-box model $\hat y$.
What the scores must inherit is exchangeability, and this requires only
that the fitting treat the $n+1$ observations symmetrically. 

\begin{definition}[One-dimensional predictive transforms]
\label[definition]{def:one_dimensional_randomized_cpd}
At input $x$, with the calibration data held fixed, the deterministic and
randomized predictive transforms are the maps
\begin{align}
y&\longmapsto G_x(y):=F_{n+1}\bigl(S_x(y)\bigr),
\label{eq:score_induced_deterministic_transform}\\
(y,\tau)&\longmapsto G_x(y,\tau):=G_{n+1}\bigl(S_x(y),\tau\bigr).
\label{eq:score_induced_cpd}
\end{align}
\end{definition}

\begin{corollary}[Coverage of the score-induced transforms]
\label[corollary]{cor:score_induced_coverage}
Let $(X_1,Y_1),\ldots,(X_{n+1},Y_{n+1})$ be exchangeable, let $\widehat y$ be
trained on an independent sample, and assume that conditionally on that
sample the scores have no ties. Let $\tau\sim\operatorname{Uniform}(0,1)$ be
independent of the data. Then, conditionally on the training sample,
\[
G_{X_{n+1}}(Y_{n+1},\tau)\sim\nu ,
\]
and for every fixed $B\subseteq[0,1]$,
\[
\mathbb P\bigl\{Y_{n+1}\in\mathcal R_B(X_{n+1})\bigr\}=\nu(B),
\text{ for }
\mathcal R_B(X_{n+1}):=\{y\in\mathcal Y:G_{X_{n+1}}(y,\tau)\in B\}.
\]
\end{corollary}
This is \Cref{lm:randomized_PIT} applied to the scores.
Uniformity by itself does not make a function a distribution; monotonicity
supplies what is missing. If $Q$ is nondecreasing and $Q(Z)\sim\nu$, then
$Q=F^\star$: the two requirements together single out the CDF, which is why a
transform meeting both deserves to be called a predictive distribution
\citep{shen2018prediction,vovk2017nonparametric}. So to interpret the
transform as a conformal predictive system, the score must also give the
required behavior of a one-dimensional cumulative distribution function, that is,
$y\mapsto G_x(y,\tau)$ must be nondecreasing.
This condition is not automatic, and \citet[Lemma~3]{vovk2017nonparametric} give a sufficient condition on the score for it to hold. A working example is the signed-residual score $S_x(y)=y-\widehat y(x)$, which is increasing in $y$; the absolute residual is not.

%% file: subfiles/optimal_transport.tex
\section{Optimal Transport and Multivariate Calibration}
\label{sec:multivariate_outputs_vector_valued_scores}

In one-dimensional statistics, a point's rank counts how many sample
observations lie at or below it. In higher dimensions, an initial attempt is to measure each point's distance from the origin.
But a distance has no fixed probability meaning: a ball of radius two may
contain almost all the mass of one distribution and very little of another.
Moreover, spherical regions in the original space may poorly reflect a
distribution that is stretched or skewed in particular directions. Directly taking the norm discards directional information. For example, the residual vectors
$(2,2)$ and $(2,-2)$ have the same norm, although the first records errors
in the same direction and the second records opposite errors. In the
model-aggregation example, these represent a shared error and a
disagreement between the models. A threshold on the norm treats them
identically. It cannot, for example, allow larger shared errors while
being more restrictive about disagreement.
A linear transformation could separate these directions before taking a
norm. It can rescale the errors and account for correlations, producing
ellipsoidal regions in the original space. However, this remains a linear
change of coordinates and cannot generally account for a curved or
asymmetric distribution of errors. The same limitation already appears
with scalar residuals: using $|y-\widehat y(x)|$ gives an interval centred
on $\widehat y(x)$, extending equally on both sides even when the errors
are skewed toward one side.

\subsection{Optimal Transport as a Multivariate CDF}
\label{subsec:multivariate_quantile_ot}
\label{subsec:multivariate_oracle_quantiles}

Instead of measuring distance in the original data space (where
shapes and correlations can vary) optimal transport approach maps the entire data distribution
onto a simple and known reference distribution, often chosen as the uniform distribution on a
unit ball or cube. This map is found by minimizing the total transport cost, typically measured
by squared Euclidean distance. Now the key benefit is that once the data are mapped to a
uniform ball, the radial shells in that ball each contain the same fraction of the probability
mass. As a result, you can rank a point by its radius in the reference space, effectively capturing
how much of the data distribution lies inside that radius. These center-outward ranks then
act like univariate ranks-completely distribution-free, because they depend only on proportions
of mass, not the specific geometry of the data. These ideas was developped in \citet{chernozhukov2017} and \citet{hallin2021}.
Nevertheless, many maps meet it, and each would give a
calibrated rank, so matching alone does not single out a notion of rank.
The cost we ask the map to minimize does. On the line, ``increasing''
reads
\[
\big[z\leq z'\Rightarrow F^\star(z)\leq F^\star(z')\big]
\Longleftrightarrow 
\bigl(F^\star(z)-F^\star(z')\bigr)(z-z')\geq0,
\]
and this asks only for a product of differences to be nonnegative and so the condition survives in $\mathbb R^d$ as
\[
\langle T(z)-T(z'),z-z'\rangle\geq0,
\]
which is what it means for $T$ to be a \emph{monotone operator}
\citep{bauschke2017}. Gradients of convex functions are monotone in this
sense, and by Brenier's theorem \citep{Bre91} the squared Euclidean cost
selects exactly such a gradient. Monotonicity thus makes $T^\star$ a
distribution function, which is
why the assignment we build in \Cref{sec:multivariate_dempster_hill} has
to be shown to inherit it \citep[\Cref{prop:monotonicity_assignment};
see also][Section~3]{hallin2021}.

\begin{definition}[Center-outward distribution function and quantile regions]
\label[definition]{def:center_outward}
Let $Z\sim\mu$ have an absolutely continuous law on $\mathbb R^d$ with
finite second moment, and let $\nu$ be the spherical-uniform law on the
unit ball $\mathcal U$. The \emph{center-outward distribution function} of
$\mu$ solves the quadratic transport problem
\begin{equation}
\label{eq:brenier_map}
T^\star\in\argmin_{T_\sharp\mu=\nu}
\int_{\mathbb R^d}\|z-T(z)\|^2\,d\mu(z),
\end{equation}
and for $r\in[0,1]$ the \emph{center-outward quantile region} of order $r$
is the pullback of the centered ball
$B(0,r)=\{u\in\mathcal U:\|u\|\leq r\}$,
\begin{equation}
\label{eq:population_center_outward_region}
\mathcal Q_{T^\star}(r):=(T^\star)^{-1}\bigl(B(0,r)\bigr)
=\bigl\{z\in\mathbb R^d:\|T^\star(z)\|\leq r\bigr\},
\end{equation}
the same pullback notation used for $F^\star$ in
\Cref{subsec:from_ground_truth_to_empirical_prediction_sets}.
\end{definition}

The constraint $T_\sharp\mu=\nu$ says that $X \sim\mu$, implies $T(X)\sim\nu$. The next statement collects the properties of $T^\star$ that the rest of this
paper uses, all of them established by \citet{hallin2021} for the
construction introduced by \citet{chernozhukov2017}. Their center-outward
distribution function $\mathbf F_\pm$ is our $T^\star$, and their
center-outward quantile region $\mathbb C(r)$ is our
$\mathcal Q_{T^\star}(r)$.
In the following, we consider the spherical-uniform
distribution on the unit ball $B(0,1)$; its density on $\mathbb{B}(0,1)\setminus\{0\}$ is
\[
f(\mathbf{u})
= \frac{\mathds{1}_{\mathbb{B}(0,1)\setminus\{0\}}(\mathbf{u})}{c_d}
\times \frac{1}{\|\mathbf{u}\|^{d-1}},
\qquad
c_d = \frac{2\pi^{d/2}}{\Gamma(d/2)} = |\mathbb{S}^{d-1}|,
\]
where $c_d$ is the surface area of the unit sphere, ensuring that the total
probability mass integrates to one. Equivalently, this is the distribution for which the radius is uniform on $[0,1]$ which is why we use the following density.

\begin{theorem}[Center-outward calibration]
\label[theorem]{thm:center_outward_calibration}
With $T^\star$ and $\mathcal Q_{T^\star}(r)$ as in \Cref{def:center_outward}:
\begin{enumerate}[label=(\roman*)]
\item $T^\star=\nabla\phi$ for a convex $\phi$, and is unique
$\mu$-almost everywhere;
\item $T^\star(Z)\sim\nu$, so $T^\star$ is a probability integral
transform and characterizes $\mu$;
\item writing the \emph{rank} and \emph{sign} of $Z$ as
\begin{equation}
\label{eq:population_uniform_rank}
\operatorname{Rank}(Z):=\|T^\star(Z)\|,
\qquad
\operatorname{Sign}(Z):=T^\star(Z)/\|T^\star(Z)\|,
\end{equation}
these are independent, with
$\operatorname{Rank}(Z)\sim\operatorname{Uniform}[0,1]$ and
$\operatorname{Sign}(Z)$ uniform on the unit sphere;
\item $\mu(\mathcal Q_{T^\star}(r))=r$ for every $r\in[0,1]$, and the regions are
nested, so $\operatorname{Rank}(z)$ is the mass of the smallest one
containing $z$;
\item for $d=1$ the unit ball is $[-1,1]$, $T^\star=2F^\star-1$, and
$\mathcal Q_{T^\star}(r)=[q_{(1-r)/2},\,q_{(1+r)/2}]$.
\end{enumerate}
\end{theorem}

For the spherical-uniform reference, a ball of radius $r$ contains
probability $r$. To obtain coverage $1-\alpha$, we simply retain the
source values whose images lie within radius $1-\alpha$. This gives the
quantile region $\mathcal Q_{T^\star}(1-\alpha)$. Although the reference
region is a ball, the corresponding region in the original space can
reflect the asymmetry and dependence in the source distribution.
In one dimension, this construction gives the familiar central interval
$[q_{\alpha/2},q_{1-\alpha/2}]$ from
\Cref{subsec:from_ground_truth_to_empirical_prediction_sets}.
Choose a fixed
set $B$ with the desired reference probability, then collect the source
values that the map sends into $B$. This reads
\begin{align}
\mathcal Q_{T^\star}(B)
&=\{z\in\mathbb R^d:T^\star(z)\in B\}=(T^\star)^{-1}(B),
\label{eq:multivariate_pullback}\\
\mathbb P\{Z\in\mathcal Q_{T^\star}(B)\}
&=\mathbb P\{T^\star(Z)\in B\}=\nu(B).
\label{eq:PIT_ground_truth}
\end{align}
Thus $\nu(B)\geq1-\alpha$ gives coverage at least $1-\alpha$.
The remaining difficulty is that the source
law $\mu$, and hence the population transport $T^\star$, are unknown.

\begin{remark}[Beyond square-integrability]
\label[remark]{rmk:no_moments}
We keep the quadratic cost throughout, since it is the cost the empirical
assignment will minimize, and it requires finite second moments. The
objects themselves need less: following \citet{delbarrio2020},
\citet{hallin2021} construct $T^\star$ as the gradient of a convex
function pushing $\mu$ to $\nu$ under no moment assumption, and obtain the
corresponding Glivenko--Cantelli result
\citep{glivenko1933,cantelli1933} without compact support or second
moments. The
regularity under which the regions $\mathcal Q_{T^\star}(r)$ are closed, connected
and strictly nested with continuous boundaries is given there as well.
\end{remark}

\subsection{Why Empirical Transport Is Not Enough?}
\label{subsec:empirical_transport_failure}

Given calibration observations $Z_1,\ldots,Z_n\in\mathbb R^d$ and fixed
reference points $U_1,\ldots,U_n\in\mathcal U$, let $\mu_n$ be their
empirical source law as in \Cref{eq:empirical_cdf}, and let
$\nu_n=n^{-1}\sum_{j=1}^n\delta_{U_j}$. Across sample sizes, choose the
reference grids so that $\nu_n\Rightarrow\nu$. The empirical transport
problem assigns each observation to a different reference point:
\begin{equation}
\label{eq:empirical_ot_assignment}
\sigma_n\in\argmin_{\sigma\in\mathfrak S_n}
\sum_{i=1}^n\|Z_i-U_{\sigma(i)}\|^2.
\end{equation}
It defines
\begin{equation}
\label{eq:empirical_ot_map_atoms}
T_n(Z_i)=U_{\sigma_n(i)},\qquad i\in[n],
\end{equation}
so
$(T_n)_\sharp\mu_n=\nu_n$.
At $d=1$ this is the empirical distribution function. Expanding the square
in \Cref{eq:empirical_ot_assignment} gives
\[
\sum_{i=1}^n\bigl(Z_i-U_{\sigma(i)}\bigr)^2
=\sum_{i=1}^nZ_i^2+\sum_{j=1}^nU_j^2-2\sum_{i=1}^nZ_iU_{\sigma(i)},
\]
and neither of the first two sums depends on $\sigma$: the observations
are untouched by it, and a permutation only reorders the reference points
it sums over. Minimizing the assignment cost is therefore the same as
maximizing $\sum_iZ_iU_{\sigma(i)}$, and on the line the rearrangement
inequality solves that by sorting: $\sigma_n$ matches the $i$-th smallest
observation to the $i$-th smallest reference point. The expansion holds in
any dimension, with $Z_iU_{\sigma(i)}$ replaced by
$\langle Z_i,U_{\sigma(i)}\rangle$; what belongs to $d=1$ is that sorting
then solves it. On $\mathcal U=[-1,1]$, with the grid
$U_{(j)}=2j/n-1$, this reads
\[
T_n(Z_{(i)})=\frac{2i}{n}-1=2F_n(Z_{(i)})-1,
\]
the empirical counterpart of $T^\star=2F^\star-1$ in
\Cref{thm:center_outward_calibration}(v). In one dimension the assignment
problem therefore returns the empirical
distribution function of \Cref{subsec:dempster_hill}.

A Pain point is that the assignment defines the map only at the calibration observations. So to apply the map to a future observation we must
choose an extension $\overline T_n:\mathbb R^d\to\mathcal U$; see, for
example, \citet[Smooth interpolation in Section 3]{delbarrio2020}. We can
then form the plug-in region
$\mathcal Q_{\overline T_n}(B)=\overline T_n^{-1}(B)$. For a central ball,
\begin{equation}
\label{eq:empirical_ot_plugin_region}
\mathcal Q_{1-\alpha,n}
=\overline T_n^{-1}(B(0,1-\alpha))
=\{z\in\mathbb R^d:\|\overline T_n(z)\|\leq1-\alpha\}.
\end{equation}

Consistency results for empirical transport
\citep[Theorem~4.1]{ghosal2021multivariateranksquantilesusing}
\citep[Proposition~3.3]{delbarrio2020} justify population approximation
under their regularity and reference-grid conditions. The plug-in argument
needs more than consistency: it needs a rate, and one good enough to pay
for widening the reference region. On the line the
Dvoretzky--Kiefer--Wolfowitz inequality supplies exactly that, bounding
$\sup_t|F_n(t)-F^\star(t)|$ by $\sqrt{\log(2/\delta)/2n}$ with probability
$1-\delta$, for every $\mu$, with explicit constants carries this through.
Rates for transport maps exist as well, and they are worse in three ways.
\citet[Theorem~6]{hutter2021minimax} show that over maps of smoothness
$\alpha\geq1$ no estimator can do better than
\begin{equation}
\label{eq:transport_minimax_rate}
\inf_{\widehat T}\ \sup_{T^\star}\
\mathbb E\int\bigl\|\widehat T-T^\star\bigr\|^2\,d\mu
\ \gtrsim\ n^{-\frac{2\alpha}{2\alpha-2+d}},
\end{equation}
and give an estimator attaining it up to logarithmic factors. At
$\alpha=1$, the smoothness Brenier's theorem already supplies, the right
side is $n^{-2/d}$: an $L^2(\mu)$ error of order $n^{-1/d}$, so certifying
error $\epsilon$ costs $n\gtrsim\epsilon^{-d}$, exponential in the
dimension; See
\citet{chewi2024statistical} for a survey.

The scalar construction in \Cref{subsec:from_ground_truth_to_empirical_prediction_sets}
suggests how to obtain marginal calibration directly. Include the future
observation in the transport problem itself. Since it is unknown, assign
each candidate $z$ using the augmented source measure $\mu_{\zeta(z)}$
from \Cref{eq:augmented_empirical_cdf}. At $z=Z_{n+1}$, this becomes
$\mu_{n+1}$, and all $n+1$ exchangeable observations enter the same
assignment. This symmetry will give the exact reference-label law in
\Cref{thm:reference_label_calibration}. Transport approximation remains
relevant to the geometry and efficiency of the resulting regions.

%% file: subfiles/mcpd.tex
\section{Multivariate Dempster--Hill Construction}
\label{sec:multivariate_dempster_hill}
\Cref{subsec:empirical_transport_failure} left the plug-in quantile region
without a guarantee, for two reasons: the empirical map says nothing at a
new point, and controlling the error of any extension of it would need a
rate that is in the wrong norm and exponential in $d$. Putting the
candidate into the transport problem itself removes both at once. In one
dimension this is exactly what augmentation did, giving the candidate a
rank and hence one of the equal-mass reference intervals; we now do the
same for vectors. We partition the reference space into $n+1$ equal-mass cells,
assign the augmented observations to these cells, and retain candidates
whose labels belong to a chosen collection. Exchangeability will determine
the coverage of this rule. Each label also identifies a cell in source
space. As in the scalar Dempster--Hill construction, choosing how to
distribute probability inside these cells gives a predictive law.

\subsection{Augmented Optimal Transport}
\label{subsec:augmented_optimal_transport}

The obstacle of \Cref{subsec:empirical_transport_failure} was that the
candidate arrives after the transport has been fitted. We remove it by
letting the candidate enter the transport problem itself, so that each
candidate carries its own problem. For simplicity, we mostly focus on the matching with the squared Euclidean cost
$c(a,u):=\|a-u\|^2$ but the results are clearly not restricted to it.

\begin{definition}[Augmented optimal transport]
\label[definition]{def:augmented_transport}
Let $u_1,\ldots,u_{n+1}$ be target points in the reference space
$\mathcal U$, not necessarily distinct, fixed independently of the
observations. For a candidate $z\in\mathbb R^d$, we recall that $\zeta(z)=(Z_1,\ldots,Z_n,z)$ be the
augmented score vector of
\Cref{eq:candidate_augmented_score_vector} and
the augmented source law to be send to the fixed target
\begin{equation}
\mu_z
=
\frac1{n+1}\Bigl(\sum_{i=1}^n\delta_{Z_i}+\delta_z\Bigr) \text{ and }
\nu_{n+1}
=
\frac1{n+1}\sum_{j=1}^{n+1}\delta_{u_j}.
\end{equation}
Both laws carry $n+1$ atoms of equal mass, and
we call \emph{augmented optimal transport at $z$} the assignment problem
between them,
\begin{equation}
\label{eq:candidate_augmented_assignment}
\sigma_z
:=
\sigma_{\zeta(z)}
\in
\argmin_{\sigma\in\mathfrak S_{n+1}}
\sum_{i=1}^{n+1}c\bigl(\zeta_i(z),u_{\sigma(i)}\bigr).
\end{equation}
Its solution induces the map
\begin{equation}
\label{eq:full_augmented_transport}
T_{n+1}^{z}\bigl(\zeta_i(z)\bigr):=u_{\sigma_z(i)},
\qquad i\in[n+1],
\end{equation}
on the augmented sample, and the label and image of the candidate are
\begin{equation}
\label{eq:candidate_label}
k(z):=\sigma_z(n+1),
\qquad
T(z):=T_{n+1}^{z}(z)=u_{k(z)}.
\end{equation}
\end{definition}

There is one such problem for every $z\in\mathbb R^d$, hence a whole
family of maps $\{T_{n+1}^{z}\}_{z\in\mathbb R^d}$ rather than a single
one: the source law $\mu_z$ moves with the candidate, so the whole
assignment may change with it. Each member of the family is exact, since
every target is used exactly once and
\begin{equation}
\label{eq:full_augmented_transport_pushforward}
(T_{n+1}^{z})_\sharp\mu_z=\nu_{n+1}
\qquad\text{for every }z\in\mathbb R^d .
\end{equation}
This removes the two obstacles of
\Cref{subsec:empirical_transport_failure}.
It also restores a symmetry that $T_n$ had broken. The empirical map is
built from the $n$ calibration scores and the candidate is only evaluated
on it, so the two are not treated alike. In
\Cref{eq:candidate_augmented_assignment} all $n+1$ points enter the same
problem in the same way. At $z=Z_{n+1}$ the candidate is one of the $n+1$
observations, and nothing in the problem distinguishes it from the others.\looseness=-1

\paragraph{A discrete reference grid}
Following \citet{hallin2021}, we discretize the spherical uniform distribution and split the
budget as
$
n+1=n_Rn_S+n_0,$ with
$0\leq n_0<n_S$,
and place
\begin{itemize}
\item $n_0$ points at the origin;
\item $n_S$ unit vectors $\theta_1,\ldots,\theta_{n_S}$ spread over the
unit sphere;
\item $n_R$ radii, regularly spaced as
$\tfrac1{n_R},\tfrac2{n_R},\ldots,1$.
\end{itemize}
The targets are the products $r\theta_\ell$ over these radii and
directions, together with the copies of the origin: concentric shells,
each carrying $n_S$ points along the chosen directions. Every target
receives mass $1/(n+1)$, so the origin carries $n_0/(n+1)$ and each
nonzero radius $j/n_R$, shared by $n_S$ points, carries $n_S/(n+1)$.

\paragraph{Equal-Mass Reference Cells}

In the scalar construction, each possible rank selects an interval with
reference mass $1/(n+1)$. We need $n+1$ reference cells, each carrying
the probability of one label. Fix their partition independently of the
calibration and future observations:
\begin{equation}
\label{eq:equal_mass_reference_cells}
\mathcal U
=
\bigsqcup_{j=1}^{n+1}L_j,
\qquad
\nu(L_j)
=
\frac{1}{n+1}.
\end{equation}

\begin{remark}[Equal-mass reference cells]
Semi-discrete optimal transport gives another way to construct the
reference cells. Fix $n+1$ sites and transport $\nu$ to them, giving each
site mass $1/(n+1)$. The points sent to each site form a Laguerre cell.
Optimizing the dual weights adjusts the cell boundaries so that each
cell has the required mass
\citep{aurenhammer1998minkowski,merigot2011multiscale,kitagawa2019convergence,levy2025large}.
An earlier version of this paper followed this strategy. Here we use an
explicit radial--directional partition, which gives equal-mass cells
without solving a semi-discrete transport problem. We leave the study of
how alternative partitions affect the prediction regions for future work.
\end{remark}

\begin{figure}[t]
\centering
\paperfigure{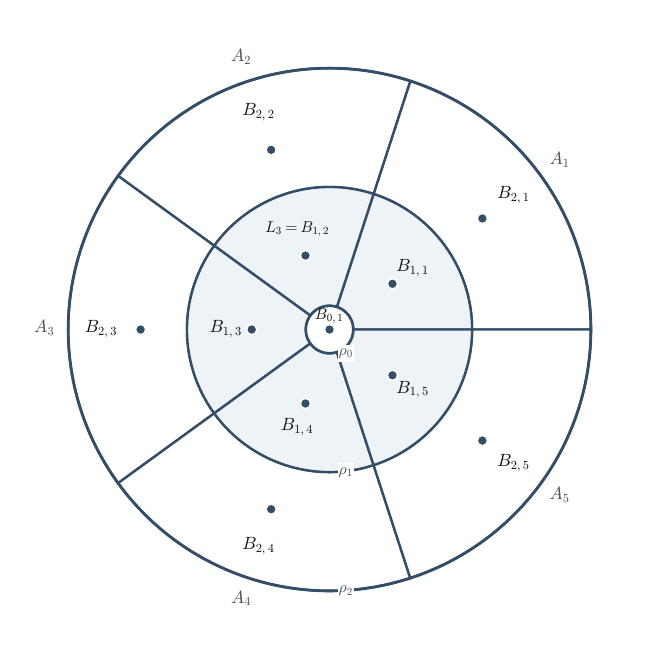}
\caption{An equal-mass reference partition with
$n+1=11=1+2\cdot5$. The central remainder and the two radial levels crossed
with five directions form eleven cells of equal $\nu$-mass. Their Euclidean
areas differ because $\nu$ is uniform in radius.}
\label{fig:radial_directional_reference_cells}
\end{figure}

For the spherical-uniform law, one possible
partition uses angular sectors and radial intervals;
\Cref{fig:radial_directional_reference_cells} gives an example. 
Here is one way to construct the equal-mass cells. It divides the
reference ball by radius and direction, as in
\Cref{fig:radial_directional_reference_cells}. 

\begin{lemma}[Equal-mass radial--directional cells]
\label[lemma]{lem:equal_mass_radial_directional_cells}
Let
\[
U=R\Theta\sim\nu,
\qquad
R\sim\operatorname{Uniform}[0,1],
\qquad
R\perp\!\!\!\perp\Theta,
\]
where $\Theta$ has uniform law $\omega$ on $\mathbb S^{d-1}$. Let
$A_1,\ldots,A_{n_S}$ partition $\mathbb S^{d-1}$ with
$\omega(A_\ell)=1/n_S$,
and let $n_R,n_0$ as before.
For $h=1,\ldots,n_0$, define the
central cells, with none when $n_0=0$,
\begin{equation}
\label{eq:central_reference_cells_main}
B_{0,h}
:=
\left\{
r\theta:
\frac{h-1}{n+1}\leq r<\frac{h}{n+1},\;
\theta\in\mathbb S^{d-1}
\right\}.
\end{equation}
For the remaining radial levels, set
\[
\rho_k
:=
\frac{n_0+kn_S}{n+1},
\qquad
k=0,\ldots,n_R,
\]
and define
\begin{equation}
\label{eq:radial_directional_reference_cells}
B_{k,\ell}
:=
\left\{
r\theta:
r\in[\rho_{k-1},\rho_k),\;
\theta\in A_\ell
\right\}.
\end{equation}
Then the $n_0$ central cells and the $n_Rn_S$ radial--directional cells
partition $\mathcal U$ up to $\nu$-null boundaries, and every cell has
$\nu$-mass $1/(n+1)$.
\end{lemma}

The pair-indexed cells are flattened into a single list by
\begin{equation}
\label{eq:reference_cell_flattening}
L_h:=B_{0,h},
\qquad
L_{n_0+(k-1)n_S+\ell}:=B_{k,\ell}.
\end{equation}

As in dimension one, the randomization within the $j$th cell uses the conditional law
$\nu(\,\cdot\mid L_j)$. 
To assign an observation to a cell,
we will average its squared distance to a draw from that cell. The cell
barycenter and its mean squared radius are the two quantities needed to
compute this cost:
\begin{equation}
\label{eq:reference_cell_moments}
m_j
=
\int_{L_j}u\,\nu(du\mid L_j),
\qquad
q_j
=
\int_{L_j}\|u\|^2\,\nu(du\mid L_j).
\end{equation}
The barycenters define the discrete reference measure
\begin{equation}
\label{eq:discrete_barycentric_reference_measure}
\nu_{n+1}
=
\frac{1}{n+1}
\sum_{j=1}^{n+1}
\delta_{m_j}.
\end{equation}

We choose cells with pairwise distinct barycenters. Randomization uses the
conditional reference laws $\nu(\,\cdot\mid L_j)$; the assignment itself
will depend only on the barycenters.
We now make the choice announced above and take the targets of
\Cref{def:augmented_transport} to be the cell barycenters,
\begin{equation}
\label{eq:barycenters_as_targets}
u_j:=m_j,
\qquad j\in[n+1],
\end{equation}
so that the target law $\nu_{n+1}$ is the one in
\Cref{eq:discrete_barycentric_reference_measure}.
This choice keeps the assignment problem of
\Cref{def:augmented_transport} and gains the cells. Sending a source point
into a whole cell rather than onto a single point is a different problem,
with cost
\begin{equation}
\label{eq:reference_cell_cost}
\begin{aligned}
C_{ij}(z)
&=
\int_{L_j}
c\bigl(\zeta_i(z),u\bigr)
\,\nu(du\mid L_j)
=
\|\zeta_i(z)\|^2
-
2\langle \zeta_i(z),m_j\rangle
+
q_j,
\end{aligned}
\end{equation}
while \Cref{eq:candidate_augmented_assignment} uses the barycentric
entries
\begin{equation}
\label{eq:barycentric_assignment_cost}
\bar C_{ij}(z)
=
c\bigl(\zeta_i(z),m_j\bigr)
=
\|\zeta_i(z)-m_j\|^2 .
\end{equation}
The first term depends on the whole conditional law inside every cell while the
second only on its mean. So both are minimized by the same permutations
(\Cref{prop:cell_barycentric_equivalence}). One solve on $n+1$ points, as
on the line, therefore returns the label $k(z)$, and the conditional laws
$\nu(\,\cdot\mid L_j)$ are still there to randomize inside the selected
cell, at no further solve.

\begin{definition}[Randomized reference kernel]
\label[definition]{def:reference_kernel}
The randomized reference kernel is the Markov kernel $K$ from
$\mathbb R^d$ to $\mathcal U$ given by
\begin{equation}
\label{eq:reference_kernel}
K(z,A)
:=
\nu\bigl(A\mid L_{k(z)}\bigr)
=
\frac{\nu\bigl(A\cap L_{k(z)}\bigr)}{\nu\bigl(L_{k(z)}\bigr)},
\qquad
A\subseteq\mathcal U\ \text{Borel},
\end{equation}
and $W_z\sim K(z,\cdot)$ denotes a draw from it.
\end{definition}

The cells have equal mass, $\nu(L_j)=1/(n+1)$ by
\Cref{eq:equal_mass_reference_cells}, so the denominator is the same for
every label and
$
K(z,A)=(n+1)\,\nu\bigl(A\cap L_{k(z)}\bigr).
$
Every $K(z,\cdot)$ is a probability measure supported in
$L_{k(z)}$, and $z\mapsto K(z,A)$ is measurable because $k$ is a
measurable map with finitely many values, so $K$ is indeed a kernel. Its
mean is the image already assigned to the candidate,
$\int_{\mathcal U}u\,K(z,du)=m_{k(z)}=T(z)$
(\Cref{eq:kernel_barycenter}).
The second object is in source space: the candidates carrying the same
label form an assigned source cell
\begin{equation}
\label{eq:assigned_source_cell}
V_j^\dagger:=\{z\in\mathbb R^d:k(z)=j\}.
\end{equation}
The nonempty assigned cells partition $\mathbb R^d$, using the boundary
convention of \Cref{def:augmented_transport}.
When the optimal assignment is unique, reordering the source points
preserves each point's reference-cell label. This permutation equivariance
is established in \Cref{app:proof_reference_label_calibration}.

\subsection{Finite-Sample Coverage Guarantee}
\label{subsubsec:finite_sample_reference_calibration}

\begin{theorem}[Finite-Sample Calibration]
\label{thm:reference_label_calibration}
Let $Z_1,\ldots,Z_n,Z_{n+1}$ be exchangeable random vectors in
$\mathbb{R}^d$ with no ties, and suppose that the optimal assignment of the
augmented sample is almost surely unique. Given the calibration
observations, let $W_z$ be a draw from $K(z,\cdot)$. Then, averaging over
the calibration observations and the future observation,
\begin{align}
k(Z_{n+1})&\sim\operatorname{Uniform}\{1,\ldots,n+1\},
\label{eq:uniform_reference_label}\\
T(Z_{n+1})&\sim\nu_{n+1},
\label{eq:discrete_barycentric_pit}\\
W_{Z_{n+1}}
&\sim\frac1{n+1}\sum_{j=1}^{n+1}\nu(\,\cdot\mid L_j)
=\nu.
\label{eq:randomized_multivariate_pit}
\end{align}
In particular, writing
$\mathcal Q_T(B)=T^{-1}(B)=\{z\in\mathbb R^d:T(z)\in B\}$ for the pullback
of a set $B\subseteq\mathcal U$, we have for every fixed $J\subseteq[n+1]$
and every fixed Borel $B$,
\begin{align}
\mathbb{P}\bigl(k(Z_{n+1})\in J\bigr)
&=\frac{|J|}{n+1},
\label{eq:label_set_probability}\\
\mathbb{P}\bigl\{Z_{n+1}\in\mathcal Q_T(B)\bigr\}
&=\nu_{n+1}(B)
=\frac{\#\{j\in[n+1]:m_j\in B\}}{n+1},
\label{eq:barycentric_pullback_coverage}\\
\mathbb{P}\bigl(W_{Z_{n+1}}\in B\bigr)
&=\nu(B).
\label{eq:randomized_pullback_coverage}
\end{align}
\end{theorem}

Since every label is equally likely at the future observation, prediction
reduces to choosing which labels to accept. The candidates with label $j$
form the source cell $V_j^\dagger$, the analogue of an order-statistic gap
on the line. For a fixed $J\subseteq[n+1]$, collect these cells:
\begin{equation}
\label{eq:score_space_label_region}
\Omega_J
=\{z\in\mathbb R^d:k(z)\in J\}
=\bigcup_{j\in J}V_j^\dagger.
\end{equation}
For the spherical-uniform reference, we can select cells from the center
outward. Let
\begin{align}
J_r
&=\{j\in[n+1]:\|m_j\|\leq r\},\qquad r\in[0,1],
\label{eq:radial_active_labels}\\
\Omega_r
&:=\Omega_{J_r}
=\{z\in\mathbb R^d:\|T(z)\|\leq r\}.
\label{eq:radial_score_region}
\end{align}

\begin{corollary}[Coverage of the Conformalized Quantile Region]
\label[corollary]{cor:conformal_radius}
Under the assumptions of \Cref{thm:reference_label_calibration}, choose
the smallest radius that includes at least
$\lceil(n+1)(1-\alpha)\rceil$ labels. Then
\begin{align}
\mathbb{P}
\left(
Z_{n+1}\in
\Omega_{r_{\alpha,n+1}}
\right)
&=
\frac{|J_{r_{\alpha,n+1}}|}{n+1}
\geq1-\alpha.
\label{eq:radial_region_coverage}
\end{align}
where
$$r_{\alpha,n+1}
=\min\{r\in[0,1]:|J_r|\geq\lceil(n+1)(1-\alpha)\rceil\}.$$
\end{corollary}

\begin{figure}[tbp]
\centering
\paperfigure{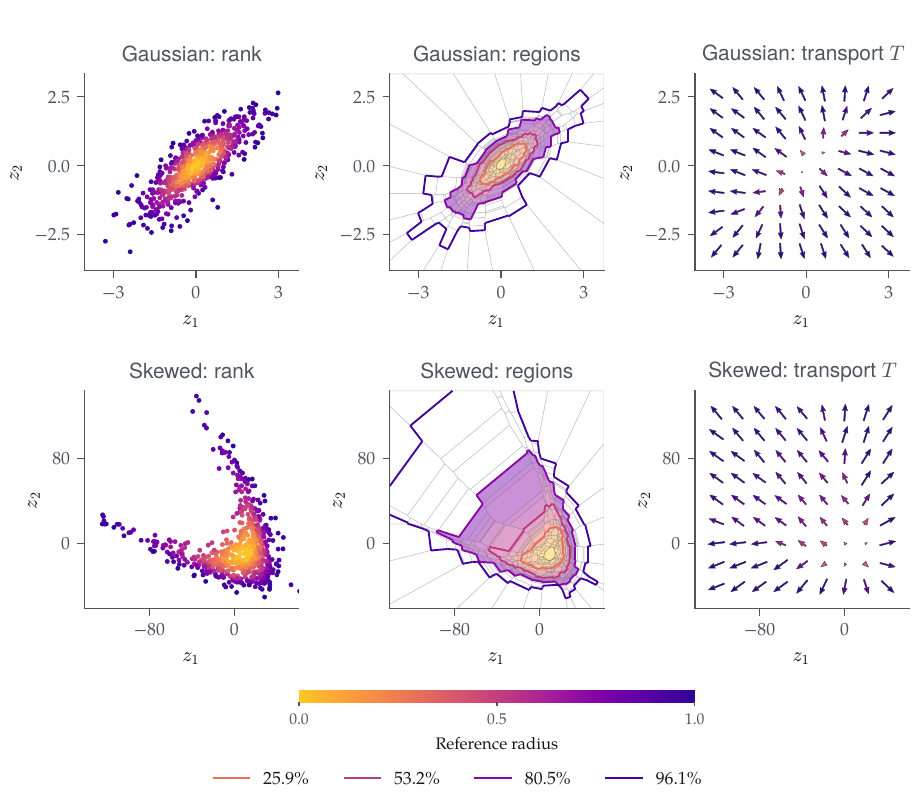}
\caption{Two samples of $n=640$ observations: a correlated Gaussian (top)
and a skewed distribution (bottom). One further draw is the candidate in the
augmented assignment. Left: observations coloured by the radius of their
assigned target. Centre: source cells and the regions $\Omega_r$ of
coverage $25.9$, $53.2$, $80.5$ and $96.1\%$; the $80.5\%$ region is filled.
Right: the map $T$. The targets form the grid of
\Cref{subsec:augmented_optimal_transport} with $n_R=n_S=25$ and $16$
copies of the origin.}
\label{fig:data_geometry}
\end{figure}

\paragraph{One-dimensional specialization.}
On the line, squared-distance matching pairs source values and reference
barycenters in increasing order: swapping a crossed pair lowers the cost.
Set $Z_{(0)}=-\infty$ and $Z_{(n+1)}=+\infty$. If
$Z_{(i)}<z<Z_{(i+1)}$, exactly $i$ calibration observations lie below $z$,
so the augmented assignment gives it label $k(z)=i+1$. Since $\nu$ is
uniform on $[-1,1]$, $K(z,\cdot)$ is uniform on the corresponding equal-mass
reference interval:
\begin{equation}
\label{eq:one_dimensional_kernel}
K(z,\cdot)
=
\operatorname{Uniform}
\left[
\frac{2i}{n+1}-1,\ \frac{2(i+1)}{n+1}-1
\right],
\end{equation}
and, with $W_z\sim K(z,\cdot)$ written as its left endpoint plus a uniform
fraction $\tau$ of its length,
\begin{equation}
\label{eq:one_dimensional_reference_pit}
\begin{aligned}
T(z)
&=\frac{2i+1}{n+1}-1
=F_{n+1}^-(z)+F_{n+1}(z)-1,\\
W_z
&=\frac{2(i+\tau)}{n+1}-1
=2G_{n+1}(z,\tau)-1,
\qquad \tau\sim\operatorname{Uniform}(0,1).
\end{aligned}
\end{equation}
The source cells are therefore the order-statistic gaps. Each contains
the future observation with marginal probability $1/(n+1)$, as in the
scalar Dempster--Hill construction. The endpoints follow the fixed
boundary convention.

\subsection{Completion of the Source Law}
\label{subsec:multivariate_dempster_hill_completion}
We now return to the question raised in the introduction: how likely is
an observation within a prediction region? The construction tells us
which source cells are accepted and gives their finite-sample coverage.
To assign probabilities within these regions, compute an expected loss,
or simulate a future observation, we need to describe how probability
is distributed inside each cell.
The kernel $K(z,\cdot)$ gives a calibrated reference coordinate for a
candidate value $z$; such a
law is obtained by reversing the correspondence between reference cells and
source cells. Each reference cell carries mass $1/(n+1)$, and the augmented
assignment associates it with one source cell. Essentially, exchangeability fixes this
equal cell mass but it does not determine how probability is distributed inside the
source cell.
\begin{definition}[Dempster--Hill Source Law]
\label[definition]{def:multivariate_dempster_hill_score_law}
Suppose that every assigned source cell $V_j^\dagger$ is nonempty. For each $j\in[n+1]$, choose a
probability law $\lambda_j$ satisfying
$
\lambda_j(V_j^\dagger)=1.
$
The corresponding source-space predictive law is
\begin{equation}
\label{eq:multivariate_dempster_hill_score_law}
\Pi
=
\frac{1}{n+1}
\sum_{j=1}^{n+1}\lambda_j.
\end{equation}
\end{definition}

This law has a direct sampling interpretation. Choose a label uniformly
from $[n+1]$, then draw a point from its filling law $\lambda_j$. Thus two
choices of filling laws can give the same probabilities to complete
cells and different probabilities to subsets within them.

For a fixed $J\subseteq[n+1]$, let
\[
V_J^\dagger
=
\bigcup_{j\in J}V_j^\dagger.
\]
Then,
\begin{equation}
\label{eq:completed_cell_mass}
\Pi(V_J^\dagger)
=
\frac{|J|}{n+1}.
\end{equation}
This is the same mass as the finite-sample marginal probability in
\Cref{eq:label_set_probability}. Therefore every admissible choice of the
within-cell laws preserves cell-aligned conformal coverage. The choice affects
moments, densities, samples, expected losses, and sets that cut through cells.
Integration and sampling formulas for $\Pi$ are discussed in the appendix.

%% file: subfiles/geometry.tex
\section{Geometry and Efficient Computation}
\label{sec:candidate_geometry_computation}
Only the candidate changes. The calibration observations and reference cells stay fixed. We use this to obtain the solution for every candidate from a single assignment solve. First, matching observations to reference cells gives the same optimal assignment as matching them to the cells’ barycenters. Indeed, replacing cell \(j\) by its barycenter subtracts \(q_j-\|m_j\|^2\) from its assignment cost, regardless of which observation is assigned to it. Every assignment uses each cell once, so its total cost decreases by the same amount. The minimizing assignment therefore remains unchanged.

\begin{proposition}[One Assignment for Cell and Barycentric Costs]
\label[proposition]{prop:cell_barycentric_equivalence}
For every candidate $z\in\mathbb{R}^d$,
\[
\argmin_{\sigma\in\mathfrak S_{n+1}}
\sum_{i=1}^{n+1}
C_{i,\sigma(i)}(z)
=
\argmin_{\sigma\in\mathfrak S_{n+1}}
\sum_{i=1}^{n+1}
\bar C_{i,\sigma(i)}(z).
\]
\end{proposition}

A direct implementation would solve an assignment for every candidate
$z\in\mathbb R^d$. There are infinitely many such candidates. Restricting
them to a finite grid does not resolve the problem completely: returning
only accepted grid points changes the prediction region and does not
preserve its coverage guarantee automatically. Moreover,
a direct grid search still fits one assignment per grid point. A grid with
$q$ locations in each of $d$ coordinates has $q^d$ points.
A similar computational difficulty arises in full conformal prediction,
where the prediction model is refitted for each candidate outcome
\citep{chen2016trimmed,johnstone2022exact,ndiaye2019computing,ndiaye2022stable,martinez2023approximating}.
Here the predictor is fixed, but the augmented assignment changes. Existing
algorithms can update a matching more cheaply than solving it again:
\citet{toroslu2007incremental} obtain an $O(n^2)$ update when one point is
added to each side, while \citet{millstettey2007dynamic} handle $k$ changes
to assignment costs in $O(kn^2)$ operations. Changing our candidate changes
one cost row, so an update can handle one candidate in $O(n^2)$ operations.
It still has to be repeated for each candidate. To describe the region over
all of $\mathbb R^d$, we instead precompute a finite set of costs that works
for every $z$.\looseness=-1

Suppose the candidate receives label $j$. It pays $\|z-m_j\|^2$, and the
calibration observations must occupy the remaining labels. Their smallest
possible cost depends on $j$, but not on $z$.
For the fixed calibration rows, write
\[
\bar C_{ij}
=
c(Z_i,m_j)
=
\|Z_i-m_j\|^2,
\qquad
i\in[n],\quad j\in[n+1].
\]
For every reference cell $k\in[n+1]$, define the leave-one-cell value
\begin{equation}
\label{eq:leave_one_cell_value}
\ell_k
=
\min_{\pi}
\sum_{i=1}^n
\bar C_{i,\pi(i)},
\text{ where }
\pi:
[n]
\longrightarrow
[n+1]\setminus\{k\}
\end{equation}

ranges over all bijections. Thus, $\ell_k$ is the minimum cost of assigning
the calibration observations to the barycenters of all reference cells except
$L_k$. By \Cref{prop:cell_barycentric_equivalence}, these barycentric values
determine the same candidate labels as the complete cell costs.
\begin{proposition}[Leave-One-Cell Representation of the Full Assignment]
\label[proposition]{prop:leave_one_cell_representation}
For each $j\in[n+1]$, fix an optimal calibration assignment
$\pi_j:[n]\to[n+1]\setminus\{j\}$ with cost $\ell_j$.
For every candidate $z\in\mathbb R^d$, the minimum total assignment cost is
\begin{equation}
\label{eq:optimal_augmented_assignment_value}
\min_{\sigma\in\mathfrak S_{n+1}}
\left\{\sum_{i=1}^n\|Z_i-m_{\sigma(i)}\|^2
+\|z-m_{\sigma(n+1)}\|^2\right\}
=\min_{j\in[n+1]}\left\{\ell_j+\|z-m_j\|^2\right\}.
\end{equation}
The candidate label satisfies
\begin{equation}
\label{eq:leave_one_candidate_label}
k(z)
\in
\argmin_{k\in[n+1]}
\left\{
\|z-m_k\|^2+\ell_k
\right\}.
\end{equation}
Using this label, define the permutation
\begin{equation}
\label{eq:reconstructed_augmented_assignment}
\sigma_z^\star(i)
=\begin{cases}
\pi_{k(z)}(i),&i\in[n],\\
k(z),&i=n+1.
\end{cases}
\end{equation}
Then $\sigma_z^\star$ is an optimal full assignment for transport from
$\mu_z$ to $\nu_{n+1}$.
\end{proposition}

The candidate therefore pays for its own label plus the best assignment of
everyone else. Once $\ell_1,\ldots,\ell_{n+1}$ are known, this same
comparison applies to every $z\in\mathbb R^d$. Candidates receiving label
$j$ can all use the same calibration assignment $\pi_j$. Thus at most
$n+1$ fixed calibration assignments suffice to recover an optimal full
assignment for every candidate.

Taken literally, the $n+1$ costs $\ell_j$ are $n+1$ separate assignment
problems, one per removed cell, hence $O(n^4)$ operations with a cubic
solver. Updating an existing assignment for each omitted cell brings the
additional work down to $O(n^3)$, but still treats the cells one at a time.
The updates all start from the same assignment. To leave another cell
available, we move a chain of observations toward the cell that was
initially free. All these chains end at the same cell in the same graph.
\Cref{prop:leave_one_shortest_path_main} therefore obtains every $\ell_j$
from one solve and one $O(n^2)$ shortest-path search. With a cubic solver,
the assignment work for all candidates is $O(n^3)$.
\looseness=-1

\subsection{Polyhedral Partitions and Monotonicity}
\label{subsec:polyhedral_assignment_representation}
A candidate prefers label $j$ to label $k$
when $\|z-m_j\|^2+\ell_j\leq\|z-m_k\|^2+\ell_k$. 
Each comparison describes a half-space. Requiring label $j$ to be at least
as good as every other label gives their intersection.
Absorb the barycenter norms into the offsets
\begin{equation}
\label{eq:potential_offsets}
\phi_j
=
\frac{\|m_j\|^2+\ell_j}{2},
\qquad
j\in[n+1].
\end{equation}
For every reference-cell label $j$, define the closed source cell
\begin{equation}
\label{eq:source_cell_definition}
V_j
=
\left\{
z\in\mathbb{R}^d:
\|z-m_j\|^2+\ell_j
\leq
\|z-m_k\|^2+\ell_k
\quad
\text{for every }k\in[n+1]
\right\}.
\end{equation}

\begin{proposition}[Polyhedral Source Cells]
\label[proposition]{prop:polyhedral_source_cells}
For every $j\in[n+1]$,
\begin{equation}
\label{eq:source_cell_halfspaces}
V_j
=
\left\{
z\in\mathbb{R}^d:
\langle z,m_k-m_j\rangle
\leq
\phi_k-\phi_j
\quad
\text{for every }k\neq j
\right\}.
\end{equation}
In particular, every $V_j$ is a convex polyhedron.
\end{proposition}

The formula also tells us what to compute. To check whether a candidate
$z$ belongs to $V_j$, evaluate the $n$ linear inequalities in
\Cref{eq:source_cell_halfspaces}. If all of them hold, label $j$ is
optimal. The accepted region is obtained by taking the union of the cells
whose labels are retained. Thus a finite list of inequalities describes
which candidates are accepted throughout $\mathbb R^d$, including values
that were never evaluated individually.

\begin{remark}[The two source cells]
\label[remark]{rmk:source_cells}
Two families of source cells appear in the paper. The assigned cells
$V_j^\dagger=\{z:k(z)=j\}$ of \Cref{eq:assigned_source_cell} are defined
by the label map, so the tie-breaking rule of
\Cref{def:augmented_transport} makes them disjoint and they partition
$\mathbb R^d$; this is what coverage statements need. The cells $V_j$ of
\Cref{eq:source_cell_definition} are defined by the costs, so they are
closed convex polyhedra; this is what the geometry needs. They differ only
on boundaries:
\[
V_j^\dagger\subseteq V_j,
\qquad
V_j\setminus V_j^\dagger
\subseteq
\bigcup_{k\neq j}
\left\{
z:\langle z,m_k-m_j\rangle=\phi_k-\phi_j
\right\}.
\]
The right-hand side is a finite union of hyperplanes, so it has Lebesgue
measure zero and carries no probability at the future observation under
the uniqueness assumption of \Cref{thm:reference_label_calibration}. Obviously, a bounded $V_j$ has bounded $V_j^\dagger$, and the two give the
same regions up to a null set.\looseness=-1
\end{remark}

\paragraph{A common monotone map}
\label{subsec:potential_representation}
Each candidate was assigned through its own transport problem. We must
therefore check that the resulting map $z\mapsto T(z)$ has a coherent
monotone geometry.

First, we recall the following result
\begin{proposition}[\citealp{combettes2018monotone}, Theorems~1.4 and~3.1]
\label[proposition]{prop:subdifferential_cyclically_monotone}
Let $\phi:\mathbb R^d\to\mathbb R$ be convex. Then $\partial\phi$ is
maximally monotone and maximally cyclically monotone: for all
$z_1,\ldots,z_q$ with $z_{q+1}=z_1$ and all $g_r\in\partial\phi(z_r)$,
\[
\langle g_1-g_2,\,z_1-z_2\rangle\geq0
\qquad\text{and}\qquad
\sum_{r=1}^{q}\langle g_r,\,z_{r+1}-z_r\rangle\leq0 .
\]
\end{proposition}

To see that the augmented transport map preserve monotonicity, it suffices to show that it derives from a convex potential.

\begin{proposition}
\label[proposition]{prop:monotonicity_assignment}
For every $z\in\mathbb{R}^d$, let us define the convex potential and candidate label 
\begin{equation}
\label{eq:global_candidate_potential}
\Phi(z)
=
\max_{j\in[n+1]}
\left\{
\langle z,m_j\rangle-\phi_j
\right\} \text{ and }
k(z)
\in
\argmax_{j\in[n+1]}
\left\{
\langle z,m_j\rangle-\phi_j
\right\}.
\end{equation}
It holds
\begin{equation}
\label{eq:barycentric_subgradient}
T(z)
=
m_{k(z)}
\in
\partial\Phi(z).
\end{equation}
Consequently, from \Cref{prop:subdifferential_cyclically_monotone}, $T$ is cyclically monotone. Moreover,
\begin{equation}
\label{eq:kernel_barycenter}
T(z)
=
\int_{\mathcal U}u\,K(z,du).
\end{equation}
\end{proposition}

The map $T$ records the mean of the reference values available in the
selected cell. Its monotonicity compares these means across candidates.
The random draw
$
W_z\sim K(z,\cdot)
$
provides the calibrated reference-space law.
We can also ask which source points share a given reference barycenter.
Those points form $V_j$. The convex conjugate of $\Phi$ describes this
inverse relation through a subgradient, allowing a whole cell to
correspond to the same barycenter.\looseness=-1
\begin{proposition}[Conjugate Source Cells]
\label[proposition]{prop:conjugate_source_cells}
For every $j\in[n+1]$ such that $V_j$ is nonempty,
\begin{equation}
\label{eq:conjugate_source_cell}
V_j
=
\partial\Phi^*(m_j).
\end{equation}
\end{proposition}

\subsection{Optimal Algorithm}
\label{subsec:efficient_candidate_computation}
The representation above leaves $n+1$ costs to compute. Solving a separate
assignment for each omitted label would be unnecessary. We obtain them
together by first finding the cheapest way to
assign the $n$ calibration observations to the $n+1$ reference labels. One
label must remain available. To use a square assignment solver, we append a
zero-cost row to $\bar C$ and denote the resulting matrix by $\widehat C$.
The label assigned to this row is precisely the one left available to the
candidate. Write this label as $k^\star$. This single solve has cost
$\ell_{k^\star}=\min_j\ell_j$.

Now suppose we want to leave label $j$ available instead. The observation
currently assigned to $j$ must move to another label. If that label is
occupied, its observation must also move. The chain ends when an observation
reaches the available label $k^\star$:
\[
j\longrightarrow a_1\longrightarrow\cdots\longrightarrow k^\star.
\]

\begin{proposition}[Leave-One-Cell Costs as Shortest-Path Distances]
\label[proposition]{prop:leave_one_shortest_path_main}
Let $\bar C\in\mathbb R^{n\times(n+1)}$, append a zero-cost row, and fix an
optimal assignment of the resulting matrix $\widehat C$. Let $k^\star$ be
the label assigned to the zero-cost row. For each $a\neq k^\star$, let
$i(a)$ be the calibration row assigned to $a$. The displacement graph
$\mathcal G^\star$ has vertices $[n+1]$ and an edge $a\to b$ for every
$a\neq k^\star$ and $b\in[n+1]$, with weight
\[
w(a\to b)=\bar C_{i(a),b}-\bar C_{i(a),a}.
\]
Then, for every $j\in[n+1]$,
\begin{equation}
\label{eq:main_shortest_path_costs}
\ell_j-\ell_{k^\star}=\operatorname{dist}_w(j,k^\star),
\end{equation}
where $\operatorname{dist}_w$ is the shortest-path weight in $\mathcal G^\star$
and $\operatorname{dist}_w(k^\star,k^\star)=0$. The minimum is attained
by a simple path. Moving the observations along such a path gives an
optimal assignment leaving label $j$ available.
\end{proposition}

Any competing assignment differs from the initial optimum by one path
from $j$ to $k^\star$ and cycles rearranging other observations. A cycle
cannot reduce the cost, since applying it would improve the initial
optimum. Discarding cycles therefore cannot hurt, while every simple path
gives a valid reassignment. This explains why the cheapest path gives the
exact difference in costs. The full argument is given in
\Cref{thm:shortest-path-formula}.

Searching for these paths is fast when the weights are nonnegative. A
single move can lower the assignment cost, so $w$ can be negative. The
dual variables from the same solve fix the sign and leave the cheapest
paths unchanged. \Cref{sec:fast_computation} does this and states the
algorithm.

The argument uses only the assignment costs. It therefore applies to
any candidate cost vector $\mathbf r=(r_1,\ldots,r_{n+1})$, whose $j$th
entry specifies what the candidate pays for label $j$:
\begin{equation}
\label{eq:all_candidate_cost_vectors}
\min_{\sigma\in\mathfrak S_{n+1}}
\left\{
\sum_{i=1}^n\bar C_{i,\sigma(i)}+r_{\sigma(n+1)}
\right\}
=
\min_{j\in[n+1]}\{r_j+\ell_j\}.
\end{equation}
A minimizing label and its reassignment path determine an optimal
augmented assignment. In our quadratic transport construction,
$r_j=\|z-m_j\|^2$. Once the $\ell_j$ are known, the inequalities in
\Cref{eq:source_cell_halfspaces} describe every source cell $V_j$.

\paragraph{Complexity}
\label{par:candidate_family_complexity}
With a dense Jonker--Volgenant solver
\citep{jonker1987shortest}, the assignment takes $O(n^3)$ operations.
The search takes $O(n^2)$ using the implementation in
\Cref{sec:fast_computation}. Including the $O(n^2d)$ operations needed
to build $\bar C$, the total preprocessing cost is
$
O(n^2d+n^3),
$
the same order as one assignment fit.
After preprocessing, evaluating a candidate's label takes $O((n+1)d)$
operations by \Cref{eq:leave_one_candidate_label}. Recovering its full
assignment from the stored path takes $O(n)$ additional operations.
The construction can use any exact assignment solver that returns optimal
dual variables. If its cost is $\mathcal T_{\mathrm{assign}}(n+1)$, our
additional work is $O(n^2)$ once the cost matrix is available. Thus
preprocessing has the same asymptotic order as the chosen solver whenever
$\mathcal T_{\mathrm{assign}}(n+1)=\Omega(n^2)$.
The initial solve could use a faster exact matching algorithm.
For Euclidean-distance costs in the plane, \citet{gattani2023robust}
give a worst-case bound of $\widetilde O(n^2\log\Delta)$, where $\Delta$
is the ratio of the largest to the smallest distance between distinct
input points. Their expected bound of $\widetilde O(n^{7/4}\log\Delta)$
assumes that both point sets are sampled independently from the same
distribution in the unit square. They also study squared-distance costs
and higher dimensions, with different bounds. Our costs are squared
distances and our reference points are fixed, so these bounds do not
apply directly. A suitable faster solver would change the cost of the
initial assignment; the $O(n^2)$ search would remain the same.\looseness=-1

\begin{figure}[tbp]
\centering
\paperfigure{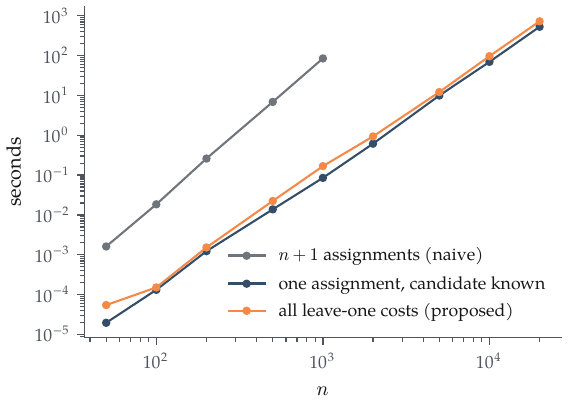}
\caption{Fitting time against $n$ for draws from the correlated Gaussian
law of \Cref{fig:data_geometry}, using barycenters of equal-mass reference
cells. One assignment of size $n+1$ with the
candidate known (reference); $n+1$ assignments of size $n$, one per omitted
label (naive); one assignment followed by the shortest-path search, which
gives every $\ell_j$ (proposed). For $n\ge100$ the proposed fit takes at
most twice as long as one assignment; the naive method takes about $n$
times longer at the tested sizes. The reference and naive fits use the
SciPy assignment solver \citep{crouse2016implementing,virtanen2020scipy};
the proposed fit uses Numba \citep{lam2015numba}. Times are from one thread
of an Apple M5 processor, with five instances for $n\le500$, three up to
$n=10^4$, and one at $n=2\cdot10^4$.}
\label{fig:timing}
\end{figure}

\subsection{Boundedness of the Quantile Region}
\label{subsec:quantile_region_boundedness}

Why can some source cells extend to infinity? Consider moving a candidate
far away in a fixed direction. In the comparison
$\|z-m_j\|^2+\ell_j$, the term $\|z\|^2$ is common to all labels.
The remaining term that grows with the candidate's distance is
$-2\langle z,m_j\rangle$. Far enough in that direction, the favored
barycenters must therefore be the ones furthest in that direction.
Such points lie on the boundary of their convex hull. A barycenter
strictly inside that hull cannot remain optimal along a ray extending
to infinity. This gives the following boundedness condition.
Let
\begin{equation}
\label{eq:barycenter_convex_hull}
\mathcal M
=
\operatorname{conv}
\left\{
m_1,\ldots,m_{n+1}
\right\}.
\end{equation}

\begin{proposition}[Boundedness of Conformal Quantile Regions]
\label[proposition]{prop:boundness_quantile_region}
If
$m_j\in\operatorname{int}(\mathcal M)$, then the source cell $V_j$ is
bounded. Consequently, the region
$\Omega_r=\bigcup_{j\in J_r}V_j^\dagger$ is bounded whenever
\[
m_j\in\operatorname{int}(\mathcal M)
\qquad\text{for every }j\in J_r.
\]
\end{proposition}

\begin{figure}[htbp]
    \centering
    \begin{subfigure}[b]{0.33\textwidth}
        \paperfigure{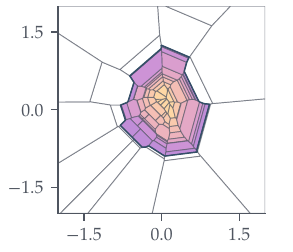}
        \caption{Radius $r=0.8$}
        \label{fig:active_cells_08}
    \end{subfigure}%
    \hfill
    \begin{subfigure}[b]{0.33\textwidth}
        \paperfigure{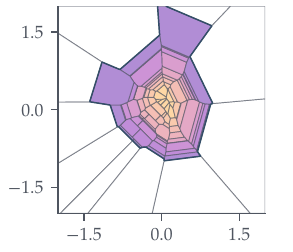}
        \caption{Radius $r=0.999$}
        \label{fig:active_cells_0999}
    \end{subfigure}%
    \hfill
    \begin{subfigure}[b]{0.33\textwidth}
        \paperfigure{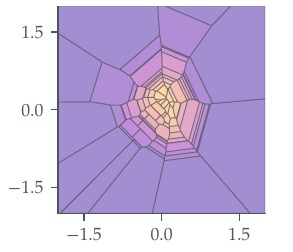}
        \caption{Radius $r=1$}
        \label{fig:active_cells_1}
    \end{subfigure}
    \caption{Nested unions for $n=99$ draws from $N_2(0,\tfrac14I_2)$ and
    $100$ targets at radii $k/10$, $k=1,\ldots,10$, along ten directions
    drawn uniformly on the circle. Retained cells are filled with the colour of
    their target's radius, as in \Cref{fig:data_geometry}; white cells are
    excluded. At $r=0.999$ the outer shell is excluded; at $r=1$ it is included.
    The common plotting window
    clips unbounded cells. For the source cells of our construction,
    \Cref{prop:boundness_quantile_region} gives a sufficient
    condition for boundedness.}
    \label{fig:conformal_cells}
\end{figure}

Boundedness is decided by where $m_j$ sits inside $\mathcal M$, and not by
how far it is from the origin. For the point grid of
\Cref{subsec:augmented_optimal_transport} the two agree, because its outer
shell has radius one: provided the directions positively span
$\mathbb R^d$, the origin is interior to $\mathcal M$, and every target of
smaller norm lies on an open segment from the origin to a vertex, hence is
interior too. Every $\Omega_r$ with $r<1$ is then bounded. Equal-mass
cells break that agreement. Their
barycenters all have norm strictly below one, so the outermost of them lie
on $\partial\mathcal M$ and their own cells are unbounded. Boundedness
then asks
$
r<\min\bigl\{\|m_j\|\ :\ m_j\in\partial\mathcal M\bigr\},
$
which excludes the outermost radial level rather than only the sphere.
\Cref{fig:conformal_cells} illustrates how a region expands as more cells
are selected.

%% file: subfiles/conformal_inference.tex
\section{Multivariate Conformal Inference}
\label{sec:supervised_prediction}
We now apply the construction to supervised prediction where one observe a sequence of features and output. The preceding
sections worked with observations $Z\in\mathbb R^d$. Given a predictor
$\widehat y$ trained on an independent sample, we take
$Z_i=S(X_i,Y_i;\widehat y)$ and write $S_x(y)=S(x,y;\widehat y)$.
At input $x$, all the operations essentially consist in composing the previous method with the score function. The label assigned to a candidate outcome $y$ is
\begin{equation}
\label{eq:output_candidate_label}
k_x(y):=k\bigl(S_x(y)\bigr).
\end{equation}
The score map transfers the source cells to the outcome space. We first
use these cells to form prediction sets, then describe predictive
transforms and laws for outcomes. We also examine when the monotonicity
of the source transform is preserved as the candidate outcome changes.\looseness=-1

\subsection{Prediction Sets}
The source partition is fixed once the calibration scores are given.
At a fixed input $x$, the score map associates each candidate outcome
$y$ with the point $S_x(y)$ in source space. The corresponding outcome
cells are the preimages
\begin{align}
V_{x,j}
&:=S_x^{-1}(V_j),
\label{eq:outcome_cell_definition}
\\
V_{x,j}^\dagger
&:=S_x^{-1}(V_j^\dagger)
=\{y\in\mathcal Y:k_x(y)=j\}.
\label{eq:assigned_outcome_cell}
\end{align}
Thus outcomes in the same assigned cell receive the same reference
label. For a fixed collection $J\subseteq[n+1]$, a candidate outcome is
retained when its score belongs to the source region, or equivalently when
its label is accepted:
\begin{equation}
\label{eq:full_transport_otcp}
\mathcal R_J(x)
=\{y\in\mathcal Y:S_x(y)\in\Omega_J\}
=\{y\in\mathcal Y:k_x(y)\in J\}.
\end{equation}
For each $x$, the nonempty assigned cells partition $\mathcal Y$,
and the prediction region is their union over the accepted labels:
\begin{equation}
\label{eq:outcome_cell_prediction_region}
\mathcal R_J(x)=\bigcup_{j\in J}V_{x,j}^\dagger.
\end{equation}
If $B_J=\bigcup_{j\in J}L_j$, a draw from
$K(S_x(y),\cdot)$ belongs to $B_J$ almost surely exactly when the label is
accepted.

\begin{proposition}[Coverage with Vector-Valued Scores]
\label[proposition]{prop:Vector_PIT_Guarantee}
Let $(X_1,Y_1),\ldots,(X_{n+1},Y_{n+1})$ be exchangeable, and suppose
that $\widehat y$ is trained on an independent sample. Assume that,
conditionally on the training sample, the scores have no ties and their
augmented optimal assignment is almost surely unique. For every fixed
$J\subseteq[n+1]$, conditionally on the training sample,
\begin{equation}
\label{eq:valid_empirical_radius}
\mathbb P\{Y_{n+1}\in\mathcal R_J(X_{n+1})\}=\frac{|J|}{n+1}.
\end{equation}
In particular, $|J|/(n+1)\geq1-\alpha$ gives coverage at least $1-\alpha$.
\end{proposition}
Selecting $k$ complete reference cells gives coverage $k/(n+1)$. To
obtain other coverage levels, choose a fixed reference set $B$ with the
desired probability. For each candidate $y$, draw a point from $\nu$
within its assigned reference cell and accept $y$ if that point belongs
to $B$. At the true future outcome, this draw has distribution $\nu$,
so the coverage is exactly $\nu(B)$.

To test whether $y$ lies in $V_{x,j}$, put $z=S_x(y)$ in
\Cref{eq:source_cell_halfspaces} and check the inequalities. The region
itself is the preimage of $\Omega_J$ under $S_x$, and for a general score
map the outcome cells can be empty or nonconvex. The residual score $S_x(y)=y-\widehat y(x)$ on
$\mathcal Y=\mathbb R^d$ is the case where everything is translated by the
fitted prediction:
\begin{equation}
\label{eq:translated_residual_region}
V_{x,j}^\dagger=\widehat y(x)+V_j^\dagger,
\qquad
\mathcal R_J(x)=\widehat y(x)+\Omega_J.
\end{equation}

The radial choice $J=J_{r_{\alpha,n+1}}$ gives the supervised prediction
set $\mathcal R_{\alpha,n+1}(x)
=\{y\in\mathcal Y:S_x(y)\in\Omega_{r_{\alpha,n+1}}\}$, which by the
definition of $r_{\alpha,n+1}$ and \Cref{prop:Vector_PIT_Guarantee}
satisfies
\[
\mathbb{P}\left(Y_{n+1}\in\mathcal R_{\alpha,n+1}(X_{n+1})\right)
\geq1-\alpha.
\]

\begin{figure}[tbp]
\centering
\paperfigure{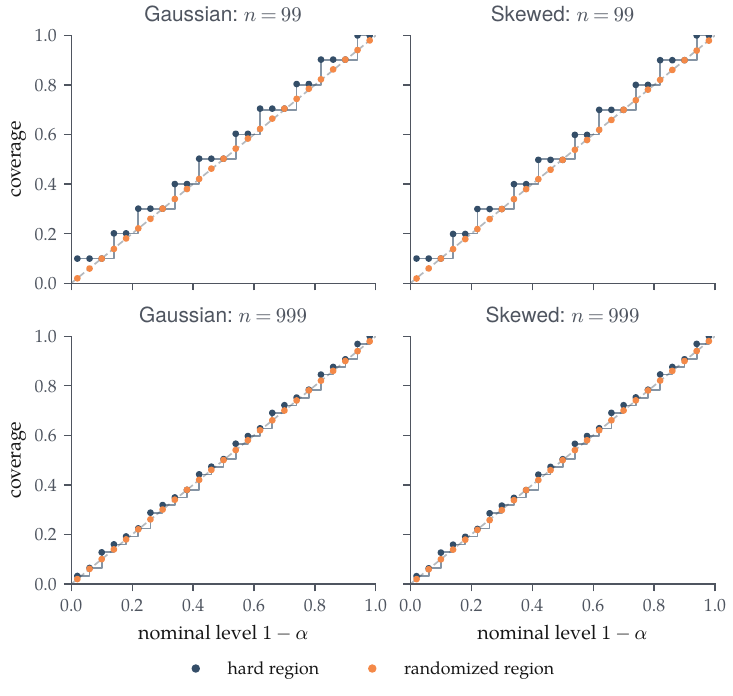}
\caption{Coverage of the prediction set against the nominal level, for
Gaussian and skewed noise, with $n=99$ (top) and $n=999$ (bottom)
calibration pairs. Each point averages $500$ repetitions with $2000$
fresh test pairs each. Targets are barycenters of equal-mass reference
cells. Hard regions follow the staircase $|J_r|/(n+1)$ and cover more than
asked; randomized-region averages lie near $1-\alpha$. The steps shrink
as $n$ grows.}
\label{fig:coverage}
\end{figure}

To illustrate the two rules, let $X$ be uniform on $[0,1]$ and
$Y=f(X)+\varepsilon$, where
$f(x)=s(\cos t(x),\sin t(x))$, $t(x)=\pi(0.9-0.8x)$, and
$\varepsilon$ follows either law in \Cref{fig:data_geometry}.
We take $s$ to be four times the largest noise standard deviation and
fit a boosted-tree predictor \citep{chen2016xgboost} on $2000$
independent pairs. \Cref{fig:coverage} compares the two rules using its
residual scores. The hard region uses the radial rule
\Cref{eq:radial_active_labels}; the randomized one draws one point in each
reference cell and keeps the cells whose draw lies in the ball of radius
$1-\alpha$.

\subsection{Predictive Distributions}
\label{subsec:multivariate_predictive_distributions}
At input $x$, the source cells become outcome cells $V_{x,j}^\dagger$.
To construct a predictive distribution, we assign each cell mass
$1/(n+1)$ and specify how to distribute that mass inside it. The
completed source law $\Pi$ provides these choices whenever we can map
its draws back to outcomes. Suppose that $S_x$ is a bijection onto an
image carrying all $\Pi$-mass. The corresponding predictive law is
\begin{equation}
\label{eq:output_predictive_law}
\Pi_x
=
(S_x^{-1})_\sharp\Pi
=
\frac1{n+1}\sum_{j=1}^{n+1}(S_x^{-1})_\sharp\lambda_j.
\end{equation}
Each term distributes probability inside the corresponding outcome
cell. To simulate an outcome, choose a label uniformly from $[n+1]$,
draw $Z'$ from its filling law $\lambda_j$, and return $S_x^{-1}(Z')$.
The assigned outcome cells therefore inherit the source cell masses:
\begin{equation}
\label{eq:completed_outcome_cell_mass}
\Pi_x(V_{x,j}^\dagger)
=\Pi(V_j^\dagger)
=\frac1{n+1}.
\end{equation}
For residual scores $S_x(y)=y-\widehat y(x)$ on
$\mathcal Y=\mathbb R^d$, this is simply a translation:
\[
Y'=\widehat y(x)+Z',
\qquad Z'\sim\Pi.
\]
Equivalently, for any measurable outcome set $A$,
\[
\Pi_x(A)=\Pi\bigl(A-\widehat y(x)\bigr).
\]
The prediction region is translated in the same way in
\Cref{eq:translated_residual_region}.

To describe a candidate outcome on the reference scale we need two
objects: a deterministic transform and a kernel that fills the cell it
selects. The requirements they must meet are the multivariate form of
those of \citet[Definition~1]{vovk2017nonparametric}.

\begin{definition}[Multivariate conformal predictive distribution]
\label[definition]{def:multivariate_cpd}
A pair $(G_x,K_x)$, with $G_x:\mathcal Y\to\mathcal U$ and $K_x(y,\cdot)$ a
probability law on $\mathcal U$, is a \emph{multivariate conformal
predictive distribution} if
\begin{enumerate}[label=(\roman*)]
\item\label{item:cpd_monotone}
for every input $x$, the map $y\mapsto G_x(y)$ is monotone,
\[
\bigl\langle G_x(y)-G_x(y'),\,y-y'\bigr\rangle\geq0
\qquad\text{for all }y,y'\in\mathcal Y;
\]
\item\label{item:cpd_calibrated}
with $W_{x,y}\sim K_x(y,\cdot)$ drawn independently of the data,
\[
G_{X_{n+1}}(Y_{n+1})\sim\nu_{n+1},
\qquad
W_{X_{n+1},Y_{n+1}}\sim\nu .
\]
\end{enumerate}
\end{definition}

Item~\ref{item:cpd_monotone} replaces monotonicity of $y\mapsto Q(y,\tau)$
and item~\ref{item:cpd_calibrated} replaces uniformity of the transform.

\begin{definition}[Predictive transform and kernel at an input]
\label[definition]{def:supervised_transform_kernel}
At input $x$, the deterministic predictive transform and the
candidate-conditional reference kernel are
\begin{align}
G_x(y)
&:=T\bigl(S_x(y)\bigr)=m_{k_x(y)},
\label{eq:multivariate_predictive_transform}\\
K_x(y,\cdot)
&:=K\bigl(S_x(y),\cdot\bigr)=\nu\bigl(\,\cdot\mid L_{k_x(y)}\bigr).
\label{eq:multivariate_cpd_kernel}
\end{align}
\end{definition}

On each assigned outcome cell $V_{x,j}^\dagger$ the transform equals $m_j$
and the kernel is $\nu(\,\cdot\mid L_j)$, so the transform is the
barycenter of the kernel,
\begin{equation}
\label{eq:supervised_kernel_barycenter}
G_x(y)
=
\int_{\mathcal U}u\,K_x(y,du).
\end{equation}
This pair meets \Cref{def:multivariate_cpd}\ref{item:cpd_calibrated} for
every score map: with $W_{x,y}\sim K_x(y,\cdot)$,
\Cref{thm:reference_label_calibration} gives it directly. It does not meet
\ref{item:cpd_monotone} for every score map, which is the subject of
\Cref{subsec:output_space_monotonicity}.

The predictive
law $\Pi_x$ distributes an unknown outcome and also depends on the
chosen filling laws $\lambda_1,\ldots,\lambda_{n+1}$. The calibration
statements average over the calibration observations and the future pair,
conditionally on the independent training sample. They do not identify
$\Pi_x$ with the true conditional law of $Y$ given $X=x$.

\subsection{Monotonicity}
\label{subsec:output_space_monotonicity}
Calibration alone does not describe how a predictive transform varies
with the candidate outcome. On the real line, a cumulative function must
also increase with that outcome. For vectors, we use the monotonicity
of the transport as the corresponding geometric property. We established
it for $T$ in score space, but the predictive transform is the composition
$G_x=T\circ S_x$. Monotonicity is not generally preserved by composition.
We therefore need to examine the score map itself.
The maps $G_x$ and $K_x$ extend the scalar construction of
\Cref{subsubsec:conformal_predictive_distributions}. In one dimension, monotonicity of the randomized map
$y\mapsto G_x(y,\tau)$ is one of the cumulative-behavior requirements for
a conformal predictive system. This requirement is studied by
\citet{vovk2017nonparametric} through candidate-dependent conformity scores.
In several dimensions, $K_x$ retains the randomized reference-space law and
$G_x$ its barycentric transform. Their calibration was established above; we
now examine how the monotonicity of the transport $T$ behaves after
composition with the score map.
Fix an input $x$ and let $S_x:\mathcal Y\to\mathbb R^d$ be a vector-valued
score map, where $\mathcal Y\subseteq\mathbb R^p$. By
\Cref{prop:monotonicity_assignment}, $T$ is monotone, and therefore
\begin{equation}
\label{eq:score_space_composed_monotonicity}
\left\langle
G_x(y)-G_x(y'),
S_x(y)-S_x(y')
\right\rangle
\geq
0.
\end{equation}
This inequality pairs increments of $G_x$ with differences in score space.
When $p=d$, monotonicity in the output variable would instead require
\begin{equation}
\label{eq:output_space_predictive_transform_monotonicity}
\left\langle
G_x(y)-G_x(y'),
y-y'
\right\rangle
\geq
0.
\end{equation}
The score difference $S_x(y)-S_x(y')$ may point in a different direction
from $y-y'$, so the first inequality does not imply the second. For
residual scores, however, these differences are equal, and monotonicity
carries over directly. The same holds when the score multiplies every
difference by a positive constant. If $p=d$ and
$
S_x(y)=\lambda y+b_x,
\text{ for }
\lambda>0,
$
then $S_x(y)-S_x(y')=\lambda(y-y')$, and
$
\lambda
\left\langle
G_x(y)-G_x(y'),
y-y'
\right\rangle
\geq
0.
$
Thus $G_x$ is monotone. The residual score
$S_x(y)=y-\widehat y(x)$ is the case $\lambda=1$.

For a general affine score
$
S_x(y)=A_xy+b_x,
\text{ where }
A_x\in\mathbb R^{d\times p},
\text{ and }
b_x\in\mathbb R^d,
$
\Cref{eq:score_space_composed_monotonicity} gives only
\[
\left\langle
G_x(y)-G_x(y'),
A_x(y-y')
\right\rangle
\geq
0.
\]
This does not generally establish the monotonicity of $G_x$ in output space,
but applying the adjoint of $A_x$ gives
\[
\begin{aligned}
&\left\langle
A_x^\top\bigl(G_x(y)-G_x(y')\bigr),
y-y'
\right\rangle=
\left\langle
G_x(y)-G_x(y'),
A_x(y-y')
\right\rangle
\geq
0.
\end{aligned}
\]
This calculation suggests how to carry the score-space geometry back to the
output space. When $S_x$ is differentiable, define
\begin{equation}
\label{eq:output_space_adjoint_pullback}
H_x(y)
=
\operatorname{Jac}_yS_x(y)^\top
G_x(y)
=
\operatorname{Jac}_yS_x(y)^\top
T\bigl(S_x(y)\bigr).
\end{equation}
When $d=p=1$, the centered midpoint identity in
\Cref{eq:one_dimensional_reference_pit} gives, away from assignment boundaries,
\[
H_x(y)
=
S_x'(y)\left[
F_{n+1}^-\bigl(S_x(y)\bigr)+F_{n+1}\bigl(S_x(y)\bigr)-1
\right].
\]
This field is generally different from the scalar randomized predictive
transform $G_x(y,\tau)$ in \Cref{eq:score_induced_cpd}.

\begin{proposition}[Output-Space Monotone Field]
\label[proposition]{prop:output_space_monotonicity}
At every $y$ where $S_x$ is differentiable and $\Phi$ is differentiable at
$S_x(y)$,
\begin{equation}
\label{eq:output_composed_gradient}
H_x(y)
=
\nabla_y
\left(
\Phi\circ S_x
\right)(y).
\end{equation}
Consequently, $H_x$ is monotone whenever this identity holds throughout
$\mathcal Y$ and $\Phi\circ S_x$ is convex.

In particular, for an affine score
$
S_x(y)
=
A_xy+b_x,
\,
A_x\in\mathbb R^{d\times p},
\text{ and }
b_x\in\mathbb R^d,
$
one has
\begin{equation}
\label{eq:affine_score_output_field}
H_x(y)
=
A_x^\top
T(A_xy+b_x),
\end{equation}
and this field is cyclically monotone. For the residual score
$
S_x(y)
=
y-\widehat y(x),
$
this reduces to $H_x(y)
=
T\bigl(y-\widehat y(x)\bigr)
=
G_x(y).
$
\end{proposition}

\begin{figure}[tbp]
\centering
\paperfigure{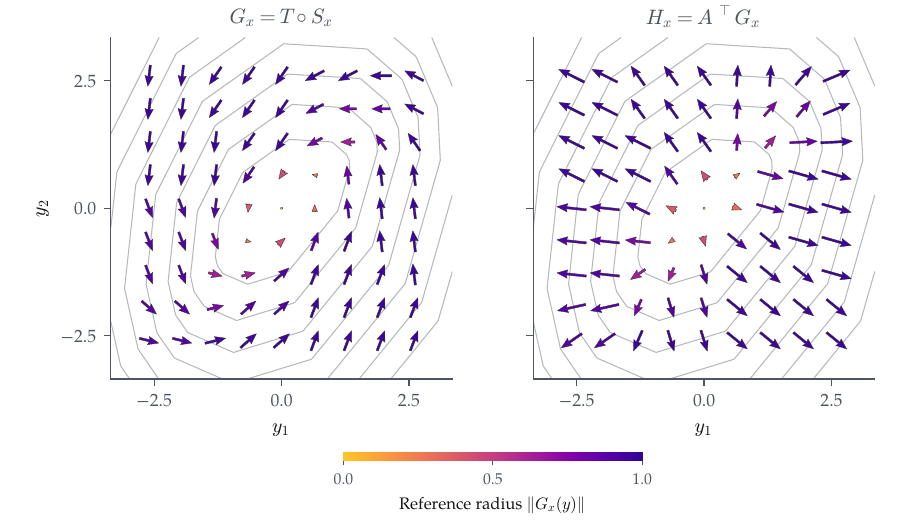}
\caption{The transform $G_x$ (left) and the field $H_x=A^\top G_x$ (right)
for an affine score. Grey lines are level sets of $\Phi\circ S_x$. $H_x$ is
its gradient and crosses them outward; $G_x$ turns along them and is not
monotone. Colour gives $\|G_x(y)\|$.}
\label{fig:predictive_fields}
\end{figure}

\Cref{fig:predictive_fields} shows both fields for the affine score
$S_x(y)=Ay$, where $A$ scales the first coordinate by $1.5$ and then
rotates by $2\pi/3$. The example uses $199$ observations from the
correlated Gaussian law of \Cref{fig:data_geometry} and $\widehat y=0$.

%% file: subfiles/proofs.tex
\section{Proofs}
\label{app:main_proofs}

\subsection{Proofs for the Multivariate Dempster--Hill Construction}
\label{app:proofs_multivariate_dempster_hill}

\subsubsection{Proof of \protect\Cref{lem:equal_mass_radial_directional_cells}}
\label{app:proof_equal_mass_radial_directional_cells}

\begin{proof}
Every nonzero $u\in\mathcal U$ has a unique polar representation
$u=r\theta$. The central radial intervals, followed by the intervals
$[\rho_{k-1},\rho_k)$, partition $[0,1]$; within every complete radial level,
$A_1,\ldots,A_{n_S}$ partition the sphere. The resulting cells therefore
partition $\mathcal U$ up to the origin and their boundaries, which have
$\nu$-probability zero.

For a central cell, the direction is unrestricted and
\[
\nu(B_{0,h})
=
\frac{h}{n+1}-\frac{h-1}{n+1}
=
\frac1{n+1}.
\]
For a radial--directional cell, polar uniqueness gives, almost surely,
\[
\{U\in B_{k,\ell}\}
=
\{R\in[\rho_{k-1},\rho_k),\;\Theta\in A_\ell\}.
\]
Independence then yields
\[
\nu(B_{k,\ell})
=
(\rho_k-\rho_{k-1})\frac1{n_S}
=
\frac{n_S}{n+1}\frac1{n_S}
=
\frac1{n+1}.
\]
\end{proof}

\subsubsection{Proof of \protect\Cref{prop:cell_barycentric_equivalence}}
\label{app:proof_cell_barycentric_equivalence}

\begin{proof}
For every $i,j\in[n+1]$,
$
C_{ij}(z)-\bar C_{ij}(z)
=
q_j-\|m_j\|^2.
$
Therefore,
\begin{align*}
\sum_{i=1}^{n+1}
\left(
C_{i,\sigma(i)}(z)
-
\bar C_{i,\sigma(i)}(z)
\right)
&=
\sum_{i=1}^{n+1}
\left(
q_{\sigma(i)}
-
\|m_{\sigma(i)}\|^2
\right)
=
\sum_{j=1}^{n+1}
\left(
q_j-\|m_j\|^2
\right),
\end{align*}
where the last equality follows because $\sigma$ is a permutation. The
difference is independent of $\sigma$, so the two costs have the same
minimizers.
\end{proof}

\subsubsection{Proof of \protect\Cref{thm:reference_label_calibration}}
\label{app:proof_reference_label_calibration}

\begin{proof}
For an ordered source vector $\mathbf z$ and
$\pi\in\mathfrak S_{n+1}$, write
$\mathbf z^\pi=(z_{\pi(1)},\ldots,z_{\pi(n+1)})$.
Reordering the sources permutes the rows of the cost matrix:
\begin{equation}
\label{eq:barycentric_cost_row_permutation}
c_{rj}(\mathbf z^\pi)
=c(z_{\pi(r)},m_j)
=c_{\pi(r),j}(\mathbf z).
\end{equation}
Thus, for any $\tau\in\mathfrak S_{n+1}$, reindexing gives
\begin{equation}
\label{eq:assignment_cost_permutation}
\sum_{r=1}^{n+1}c_{r,\tau(r)}(\mathbf z^\pi)
=\sum_{i=1}^{n+1}c_{i,(\tau\circ\pi^{-1})(i)}(\mathbf z).
\end{equation}
In particular, the sets of optimal assignments satisfy
\begin{equation}
\label{eq:assignment_argmin_permutation}
\argmin_{\tau\in\mathfrak S_{n+1}}
\sum_{r=1}^{n+1}c_{r,\tau(r)}(\mathbf z^\pi)
=
\left\{\sigma\circ\pi:
\sigma\in\argmin_{\sigma\in\mathfrak S_{n+1}}
\sum_{i=1}^{n+1}c_{i,\sigma(i)}(\mathbf z)\right\}.
\end{equation}
When the optimum is unique, this becomes
\begin{equation}
\label{eq:assignment_permutation_equivariance}
\sigma_{\mathbf z^\pi}=\sigma_{\mathbf z}\circ\pi,
\qquad
\sigma_{\mathbf z^\pi}(r)=\sigma_{\mathbf z}(\pi(r)).
\end{equation}
Each observation therefore retains its label when its position changes.
In one dimension this is the familiar fact that reordering observations
does not change their ranks.

For the augmented sample $\mathbf Z=(Z_1,\ldots,Z_{n+1})$,
exchangeability and \Cref{eq:assignment_permutation_equivariance} imply
that $\sigma_{\mathbf Z}(1),\ldots,\sigma_{\mathbf Z}(n+1)$ have the same
marginal distribution. Since each label occurs exactly once, every fixed
$J\subseteq[n+1]$ satisfies
\begin{align*}
\mathbb P\bigl(k(Z_{n+1})\in J\bigr)
&=\frac1{n+1}\sum_{i=1}^{n+1}
  \mathbb P\bigl(\sigma_{\mathbf Z}(i)\in J\bigr)\\
&=\mathbb E\!\left[
  \frac1{n+1}\sum_{i=1}^{n+1}
  \mathds{1}_{\{\sigma_{\mathbf Z}(i)\in J\}}\right]\\
&=\frac{|J|}{n+1}.
\end{align*}

Taking $J=\{j\}$ proves uniformity of the reference label. For a Borel
set $A\subseteq\mathcal U$, take instead
$J=\{j:m_j\in A\}$. Since $T(Z_{n+1})=m_{k(Z_{n+1})}$,
\[
\mathbb P\bigl(T(Z_{n+1})\in A\bigr)
=\frac{\#\{j:m_j\in A\}}{n+1}
=\nu_{n+1}(A).
\]
\end{proof}

\subsubsection{Proof of \protect\Cref{eq:randomized_multivariate_pit}}
\label{app:proof_multivariate_randomized_pit}

\begin{proof}
For every Borel set $B\subseteq\mathcal U$,
\begin{align*}
\mathbb{P}
\left(
W_{Z_{n+1}}\in B
\right)
&=
\sum_{j=1}^{n+1}
\mathbb{P}
\left(
k(Z_{n+1})=j
\right)
\nu(B\mid L_j)
=
\frac{1}{n+1}
\sum_{j=1}^{n+1}
\nu(B\mid L_j)
=
\nu(B).
\end{align*}
The second equality follows from
\Cref{thm:reference_label_calibration}, and the last one follows from the
equal-mass reference-cell partition.
\end{proof}

\subsubsection{Proof of \protect\Cref{prop:leave_one_cell_representation}}
\label{app:proof_leave_one_cell_representation}

\begin{proof}
Consider any permutation $\sigma\in\mathfrak S_{n+1}$ assigning
$Z_1,\ldots,Z_n,z$ to the reference barycenters, and put $k=\sigma(n+1)$.
Its calibration part leaves label $k$ available, so its cost is at least
$\ell_k$. Choose the smallest minimizing label
$j\in\argmin_a\{\ell_a+\|z-m_a\|^2\}$, and let $\sigma_z^\star$ extend
$\pi_j$ by assigning $z$ to $j$. Then
\begin{align*}
\sum_{i=1}^{n+1}\bar C_{i,\sigma(i)}(z)
&=\sum_{i=1}^n\|Z_i-m_{\sigma(i)}\|^2+\|z-m_k\|^2\\
&\geq\ell_k+\|z-m_k\|^2\\
&\geq\ell_j+\|z-m_j\|^2\\
&=\sum_{i=1}^{n+1}\bar C_{i,\sigma_z^\star(i)}(z).
\end{align*}
Thus $\sigma_z^\star$ is optimal, and its cost gives
\Cref{eq:optimal_augmented_assignment_value}. Conversely, an optimal
permutation must attain both inequalities, so its candidate label minimizes
$\ell_k+\|z-m_k\|^2$. The same boundary convention selects $k(z)=j$ when
several labels attain the minimum, giving the reconstruction in
\Cref{eq:reconstructed_augmented_assignment}.
\end{proof}

\subsubsection{Proof of \protect\Cref{prop:monotonicity_assignment}}
\label{app:proof_monotonicity_assignment}

\begin{proof}
The identity
$\|z-m_j\|^2+\ell_j=\|z\|^2-2(\langle z,m_j\rangle-\phi_j)$
shows that $k(z)$ selects an active affine branch of $\Phi$.
Its gradient is $m_{k(z)}$, so $T(z)\in\partial\Phi(z)$. As a maximum of
finitely many affine functions, $\Phi$ is convex, so both monotonicity
statements for $T$ follow from
\Cref{prop:subdifferential_cyclically_monotone}.

Finally, the definition of the reference barycenters gives
\[
\int_{\mathcal U}u\,K(z,du)
=\int_{L_{k(z)}}u\,\nu(du\mid L_{k(z)})
=m_{k(z)}=T(z).
\]
\end{proof}

\subsubsection{Proof of \protect\Cref{prop:conjugate_source_cells}}
\label{app:proof_conjugate_source_cells}

\begin{proof}
Choose $z_0\in V_j$. The $j$th affine branch is active at $z_0$, so
$\langle z_0,m_j\rangle-\Phi(z_0)=\phi_j$.
For every $z$, the definition of $\Phi$ gives
$\langle z,m_j\rangle-\Phi(z)\leq\phi_j$.
Consequently, $\Phi^*(m_j)=\phi_j$, and Fenchel equality gives
\begin{align*}
z\in V_j
&\quad\Longleftrightarrow\quad
\Phi(z)+\Phi^*(m_j)=\langle z,m_j\rangle\\
&\quad\Longleftrightarrow\quad m_j\in\partial\Phi(z)\\
&\quad\Longleftrightarrow\quad z\in\partial\Phi^*(m_j).
\end{align*}
\end{proof}

\subsubsection{Proof of \protect\Cref{prop:boundness_quantile_region}}
\label{app:proof_boundness_quantile_region}

\begin{proof}
The \emph{recession cone} of a nonempty closed convex set collects the
directions along which it extends to infinity,
\[
\operatorname{rec}(V_j)
=
\left\{
v\in\mathbb R^d:\ z+tv\in V_j\ \text{for all }z\in V_j,\ t\geq0
\right\},
\]
and a polyhedron is bounded exactly when its recession cone is
$\{0\}$. For a polyhedron the cone is read off its half-spaces with the
right-hand sides set to zero, so \cref{eq:source_cell_halfspaces} gives
\begin{equation}
\label{eq:source_cell_recession_cone}
\operatorname{rec}(V_j)
=
\left\{
v\in\mathbb R^d:\
\langle v,m_k-m_j\rangle\leq0
\ \text{for every }k\neq j
\right\}.
\end{equation}

Suppose $V_j$ is unbounded and take $v\in\operatorname{rec}(V_j)$ with
$v\neq0$. By \Cref{eq:source_cell_recession_cone},
$\langle v,m_k\rangle\leq\langle v,m_j\rangle$ for every $k$: the
barycenter $m_j$ maximizes the nonzero linear functional
$m\mapsto\langle v,m\rangle$ over $\{m_1,\ldots,m_{n+1}\}$, hence over
their convex hull $\mathcal M$. It therefore lies on a supporting
hyperplane of $\mathcal M$ and cannot belong to
$\operatorname{int}(\mathcal M)$.
Therefore $m_j\in\operatorname{int}(\mathcal M)$ implies that $V_j$ is
bounded. If this holds for every $j\in J_r$, the finite union of these
cells is bounded, and so is its subset $\Omega_r$.
\end{proof}

\subsubsection{Proof of \protect\Cref{prop:Vector_PIT_Guarantee}}
\label{app:proof_vector_pit_guarantee}

\begin{proof}
Conditionally on the independent training sample, the scores
$Z_i=S(X_i,Y_i;\widehat y)$, $i\in[n+1]$, are exchangeable. Moreover,
\[
Y_{n+1}\in\mathcal R_J(X_{n+1})
\quad\Longleftrightarrow\quad
k(Z_{n+1})\in J.
\]
Apply \Cref{thm:reference_label_calibration} conditionally on training
to obtain the stated coverage.
\end{proof}

\subsubsection{Proof of \protect\Cref{prop:output_space_monotonicity}}
\label{app:proof_output_space_monotonicity}

\begin{proof}
At every point where $\Phi$ is differentiable at $S_x(y)$,
\[
T(S_x(y))
=
\nabla\Phi(S_x(y)).
\]
The chain rule therefore gives
\begin{align*}
\nabla_y
\left(
\Phi\circ S_x
\right)(y)
&=
\operatorname{Jac}_yS_x(y)^\top
\nabla\Phi(S_x(y))
\\
&=
\operatorname{Jac}_yS_x(y)^\top
T(S_x(y))
\\
&=
H_x(y).
\end{align*}
The gradient of a convex function is monotone, proving the first claim.

If $S_x(y)=A_xy+b_x$, then $\Phi\circ S_x$ is convex. The affine
subgradient-chain rule gives
\[
A_x^\top T(A_xy+b_x)
\in
\partial(\Phi\circ S_x)(y),
\]
including at points where $\Phi$ is not differentiable. Hence the field is
cyclically monotone.

The residual case follows by taking $A_x=I$ and
$b_x=-\widehat y(x)$.
\end{proof}

%% file: subfiles/optimal_algorithm.tex
\section{Efficient Implementations}
\label{sec:fast_computation}
We now compute the leave-one-cell values $\ell_1,\ldots,\ell_{n+1}$ from the
calibration cost matrix $\bar C\in\mathbb R^{n\times(n+1)}$ in
\Cref{eq:leave_one_cell_value}. The derivation below applies to any finite cost
matrix; the barycentric quadratic costs enter only through the entries of
$\bar C$. Together with the fixed barycenter norms, the resulting values give
the offsets $\phi_j$ in \Cref{eq:potential_offsets}, and hence the candidate
labels and source cells.

\subsection{Augmented assignment formulation}

Define the augmented cost matrix
\[
\widehat C\in\mathbb R^{(n+1)\times(n+1)}
\]
by
\[
\widehat C_{ij}
=
\begin{cases}
\bar C_{ij}, & i\in[n],\ j\in[n+1],\\
0, & i=n+1,\ j\in[n+1].
\end{cases}
\]
The last row $i=n+1$ is called the free row.

Let $\mathcal P_{n+1}$ denote the set of $(n+1)\times(n+1)$ permutation
matrices. We consider the augmented assignment problem
\[
\min_{M\in\mathcal P_{n+1}}\langle \widehat C,M\rangle,
\qquad
\langle \widehat C,M\rangle
:=
\sum_{i=1}^{n+1}\sum_{j=1}^{n+1}\widehat C_{ij}M_{ij}.
\]

For each $k\in[n+1]$, define the fixed-free-column class
\[
\Sigma_k
:=
\{M\in\mathcal P_{n+1}:M_{n+1,k}=1\}.
\]
Thus $\Sigma_k$ is the set of augmented assignments in which the free row is
assigned to column $k$.

\begin{proposition}[Fixed-free-column values]\label[proposition]{prop:fixed-free-column-values}
\label[proposition]{app:proof_fixed_free_column_values}
For every $k\in[n+1]$,
\[
\ell_k=\min_{M\in\Sigma_k}\langle \widehat C,M\rangle.
\]
Consequently,
\[
\min_{M\in\mathcal P_{n+1}}\langle \widehat C,M\rangle
=
\min_{k\in[n+1]} \ell_k.
\]

If $M^\star$ is an optimizer of the augmented assignment problem and
\[
M^\star_{n+1,k^\star}=1,
\]
then
\[
\langle \widehat C,M^\star\rangle
=
\ell_{k^\star}
=
\min_{k\in[n+1]}\ell_k.
\]
\end{proposition}

\begin{proof}
A bijection $\pi:[n]\to[n+1]\setminus\{k\}$ extends uniquely to a
permutation matrix $M^\pi\in\Sigma_k$ by assigning the free row to
column $k$. Conversely, restricting any $M\in\Sigma_k$ to its first
$n$ rows gives such a bijection. Since the free row costs zero,
\begin{align*}
\min_{M\in\Sigma_k}\langle\widehat C,M\rangle
&=\min_{\pi:[n]\to[n+1]\setminus\{k\}\ \mathrm{bijection}}
  \left(\sum_{i=1}^n\bar C_{i,\pi(i)}+\widehat C_{n+1,k}\right)
=\ell_k.
\end{align*}
Every permutation assigns the free row to exactly one column, so the
sets $\Sigma_k$ partition $\mathcal P_{n+1}$. If $M^\star$ is an
optimum and its free column is $k^\star$, then
\begin{align*}
\min_k\ell_k
\leq\ell_{k^\star}
\leq\langle\widehat C,M^\star\rangle
=\min_{M\in\mathcal P_{n+1}}\langle\widehat C,M\rangle
 =\min_k\ell_k.
\end{align*}
All inequalities are therefore equalities.
\end{proof}

\subsection{Reference optimum and displacement graph}

Fix an optimal augmented assignment $M^\star$. Let $k^\star\in[n+1]$ be the
column assigned to the free row:
\[
M^\star_{n+1,k^\star}=1.
\]
By \Cref{prop:fixed-free-column-values},
\[
\langle \widehat C,M^\star\rangle=\ell_{k^\star}=\min_{k\in[n+1]}\ell_k.
\]

Let $\sigma^\star:[n+1]\to[n+1]$ be the permutation associated with
$M^\star$,
so that
\[
M^\star_{i,\sigma^\star(i)}=1
\qquad (i\in[n+1]).
\]
Since $\sigma^\star(n+1)=k^\star$, the real rows $1,\dots,n$ occupy exactly
the columns
\[
\mathcal A^\star:=[n+1]\setminus\{k^\star\}.
\]
For each $a\in\mathcal A^\star$, let $i(a)\in[n]$ be the unique real row occupying
column $a$ under $M^\star$:
\[
\sigma^\star(i(a))=a.
\]
The map
\[
i:\mathcal A^\star\to[n]
\]
is a bijection, because the restriction of $\sigma^\star$ to the real rows is a
bijection from $[n]$ onto $\mathcal A^\star$.

We now define the displacement graph $\mathcal G^\star$ associated with
$M^\star$. Its vertex set is $[n+1]$. For every active column
$a\in\mathcal A^\star$ and every column $b\in[n+1]$, we add a
directed edge
\[
a\to b.
\]
Thus there are no outgoing edges from $k^\star$, and every other vertex has
an edge to every vertex of $[n+1]$, including itself. In particular, every
directed cycle in $\mathcal G^\star$ is contained entirely in
$\mathcal A^\star=[n+1]\setminus\{k^\star\}$.
The weight of the edge $a\to b$ is
\[
w(a\to b)
:=
\widehat C_{i(a),b}-\widehat C_{i(a),a}
.\]
In particular,
$
w(a\to a)=0.
$
The self-loops are included only to make the notation exact: they represent
rows that do not move and contribute zero cost.

For any directed walk $W$ in $\mathcal G^\star$, we write
$
\operatorname{wt}(W)
$
for the sum of the weights of the edges traversed by $W$, counted with
multiplicity. The same notation is used for paths and cycles.

\subsection{A path-cycle lemma for partial bijections}

The next usual graph lemma is the combinatorial core of the argument.

\begin{lemma}[Path-cycle decomposition of a partial bijection]
\label[lemma]{lem:path-cycle-partial-bijection}
Let $\mathcal V$ be a finite set, and let $s,t\in\mathcal V$. Define
\[
\mathcal A:=\mathcal V\setminus\{t\},
\qquad
\mathcal B:=\mathcal V\setminus\{s\}.
\]
Suppose
\[
\rho:\mathcal A\to\mathcal B
\]
is a bijection.

Consider the directed graph $\mathcal G_\rho$ on vertex set $\mathcal V$ with one directed
edge
\[
a\to \rho(a)
\]
for every $a\in\mathcal A$. Self-loops are allowed.

Then $\mathcal G_\rho$ decomposes uniquely into one directed path from $s$ to $t$
and a disjoint collection of directed cycles, where cycles of length one are
allowed.

More precisely, its edge set can be written as a disjoint union
\[
E(\mathcal G_\rho)
=
E(P)\cup E(Q_1)\cup\cdots\cup E(Q_m),
\]
where $P$ is a directed path from $s$ to $t$, each $Q_r$ is a directed
cycle, and the vertex sets of these components are pairwise disjoint.

When $s=t$, the path $P$ is the empty path from $s$ to itself.
\end{lemma}

By extending $\rho$ to a full bijection on $\mathcal{V}$ via the mapping $t \mapsto s$, the standard cycle decomposition of permutations ensures the graph decomposes into disjoint directed cycles. Removing the edge $(t, s)$ breaks exactly one of these cycles into a simple directed path from $s$ to $t$. The uniqueness follows directly from the uniqueness of the permutation cycle decomposition

\begin{proof}
Let $E = \{(v, \rho(v)) \mid v \in \mathcal{V} \setminus \{t\}\}$ be the edge set of $\mathcal{G}_\rho$. Because $\rho : \mathcal{V} \setminus \{t\} \to \mathcal{V} \setminus \{s\}$ is a bijection, the in-degree $deg^-(v)$ and out-degree $deg^+(v)$ for every vertex $v \in \mathcal{V}$:

\begin{itemize}
\item $deg^-(s) = 0$ and $deg^+(t) = 0$.
\item For all $v \in \mathcal{V} \setminus \{s\}$, $deg^-(v) = 1$.
\item For all $v \in \mathcal{V} \setminus \{t\}$, $deg^+(v) = 1$.
\end{itemize}

Construct a walk starting at $s$ via the sequence $v_0 = s$ and $v_{k} = \rho(v_{k-1})$. Because $\mathcal{V}$ is finite, this sequence must eventually terminate or repeat. Suppose a vertex repeats, meaning $v_k = v_j$ for some $k > j \ge 0$. If $j = 0$, then $v_k = s$, which contradicts $deg^-(s) = 0$. If $j > 0$, then $v_k$ has two distinct predecessors ($v_{k-1}$ and $v_{j-1}$), which contradicts $deg^-(v) = 1$. Therefore, the sequence consists of distinct vertices and must terminate at the only vertex lacking an outgoing edge, $t$. This uniquely defines a simple directed path $P$ from $s$ to $t$.
Let $\mathcal{V}' = \mathcal{V} \setminus V(P)$. Removing $P$ from $\mathcal{G}_\rho$ eliminates exactly one incoming and one outgoing edge for every vertex in the path, leaving the degrees of the remaining vertices in $\mathcal{V}'$ unchanged. Thus, for all $v \in \mathcal{V}'$, $deg^-(v) = deg^+(v) = 1$. A finite directed graph where every vertex has an in-degree and out-degree of exactly $1$ uniquely decomposes into a set of disjoint directed cycles.
\end{proof}

\subsection{Reassignment decomposition induced by another assignment}

Fix $k\in[n+1]$, and let $M\in\Sigma_k$. Let
$\sigma_M:[n+1]\to[n+1]$ be the permutation associated with $M$:
$
M_{i,\sigma_M(i)}=1.
$
Since $M\in\Sigma_k$, we have $\sigma_M(n+1)=k$, and the restriction of
$\sigma_M$
to $[n]$ is a bijection onto $[n+1]\setminus\{k\}$.
Under $M^\star$, the real row $i(a)$ occupies column $a\in \mathcal A^\star$. Under
$M$, the same row is assigned to column $\sigma_M(i(a))$. Define
\[
\rho_M(a):=\sigma_M(i(a)),
\qquad a\in \mathcal A^\star.
\]
Also define
\[
\mathcal A_k:=[n+1]\setminus\{k\}.
\]

\begin{lemma}[Reassignment path-cycle decomposition]
\label[lemma]{lem:reassignment-decomposition}
For every $M\in\Sigma_k$, the map
$
\rho_M:\mathcal A^\star\to \mathcal A_k
$
is a bijection. The directed graph with edges
$
a\to \rho_M(a),
\, a\in \mathcal A^\star,
$

decomposes uniquely into one directed path
\[
P_M:k\to k^\star
\]
and a disjoint collection of directed cycles
\[
Q_M^{(1)},\dots,Q_M^{(m)}.
\]
If $k=k^\star$, the path $P_M$ is empty.
Moreover,
\[
\langle \widehat C,M\rangle-\langle \widehat C,M^\star\rangle
=
\operatorname{wt}(P_M)
+
\sum_{r=1}^m \operatorname{wt}(Q_M^{(r)}),
\]
where the weights are those of the displacement graph $\mathcal G^\star$.
\end{lemma}

\begin{proof}
Under $M^\star$, the map $a\mapsto i(a)$ is a bijection from the
occupied columns $\mathcal A^\star=[n+1]\setminus\{k^\star\}$ to the
calibration rows. Under $M\in\Sigma_k$, those rows occupy
$\mathcal A_k=[n+1]\setminus\{k\}$. Consequently,
$\rho_M(a)=\sigma_M(i(a))$ is a bijection from
$\mathcal A^\star$ to $\mathcal A_k$.
Apply \Cref{lem:path-cycle-partial-bijection} with $s=k$ and
$t=k^\star$. The edges $a\to\rho_M(a)$ decompose into a path
$P_M:k\to k^\star$ and disjoint cycles $Q_M^{(1)},\ldots,Q_M^{(m)}$.
The free row contributes zero to both assignments. Subtracting the
costs of the calibration rows:
\begin{align*}
\langle\widehat C,M\rangle-\langle\widehat C,M^\star\rangle
&=\sum_{a\in\mathcal A^\star}
  \bigl(\widehat C_{i(a),\rho_M(a)}-\widehat C_{i(a),a}\bigr)\\
&=\sum_{a\in\mathcal A^\star}w(a\to\rho_M(a))\\
&=\operatorname{wt}(P_M)
  +\sum_{r=1}^m\operatorname{wt}(Q_M^{(r)}).
\end{align*}
\end{proof}

\subsection{Nonnegative directed cycles}

\begin{lemma}[No negative directed cycles]
\label[lemma]{lem:no-negative-cycles}
Every directed cycle in the displacement graph $\mathcal G^\star$ has nonnegative
total weight.
\end{lemma}

\begin{proof}
Let $Q:a_1\to\cdots\to a_m\to a_1$ be a simple directed cycle, and put
$a_{m+1}=a_1$. Its vertices are occupied columns, since $k^\star$ has no
outgoing edge. Reassigning row $i(a_r)$ from $a_r$ to $a_{r+1}$ for every
$r$ produces a feasible assignment $\widetilde M\in\Sigma_{k^\star}$.
All other rows keep their columns. Optimality of $M^\star$ gives
\begin{align*}
0
&\leq\langle\widehat C,\widetilde M\rangle
     -\langle\widehat C,M^\star\rangle
=\sum_{r=1}^m
  \bigl(\widehat C_{i(a_r),a_{r+1}}-\widehat C_{i(a_r),a_r}\bigr)
=\operatorname{wt}(Q).
\end{align*}
Any closed directed walk decomposes into such cycles, so its weight is
also nonnegative.
\end{proof}

\subsection{Shortest-walk representation of the leave-one-cell values}

For $k\in[n+1]$, let $\Delta_k$ denote the minimum total weight of a directed
walk from $k$ to $k^\star$ in $\mathcal G^\star$, with the convention
\[
\Delta_{k^\star}:=0.
\]
Because $\mathcal G^\star$ contains an edge $k\to k^\star$ for every
$k\neq k^\star$, such a walk always exists.

By \Cref{lem:no-negative-cycles}, every directed cycle in $\mathcal G^\star$ has
nonnegative weight. Therefore any directed walk from $k$ to $k^\star$ can
be simplified by successively removing closed subwalks; each such closed
subwalk decomposes into directed cycles, all of which have nonnegative weight.
Hence the simplification does not increase the total weight.
Consequently, the same value $\Delta_k$ is obtained by minimizing over
simple directed paths from $k$ to $k^\star$, and the minimum is attained
because there are only finitely many such paths.

\begin{theorem}[Shortest-path formula]
\label[theorem]{thm:shortest-path-formula}
For every $k\in[n+1]$,
\[
\ell_k-\ell_{k^\star}
=
\Delta_k,
\]
where $\Delta_k$ is the shortest-walk distance from $k$ to $k^\star$ in
the displacement graph $\mathcal G^\star$.
Moreover, because every directed cycle in $\mathcal G^\star$ has nonnegative weight,
the same distance is obtained by minimizing over simple directed paths from
$k$ to $k^\star$.
\end{theorem}

\begin{proof}
Fix $k\in[n+1]$. Every $M\in\Sigma_k$ decomposes as in
\Cref{lem:reassignment-decomposition}. Since its cycles have
nonnegative weight by \Cref{lem:no-negative-cycles},
\begin{align*}
\langle\widehat C,M\rangle-\ell_{k^\star}
&=\operatorname{wt}(P_M)
  +\sum_r\operatorname{wt}(Q_M^{(r)})
\geq\operatorname{wt}(P_M)
\geq\Delta_k.
\end{align*}
Minimizing over $M\in\Sigma_k$ gives
$\ell_k-\ell_{k^\star}\geq\Delta_k$.

There is a path from $k$ to $k^\star$: the direct edge suffices when
$k\neq k^\star$. Removing cycles cannot increase its weight. A shortest
path can therefore be chosen among the finitely many simple paths.
Write one as $P:k=j_0\to j_1\to\cdots\to j_t=k^\star$.
Move row $i(j_r)$ from $j_r$ to $j_{r+1}$ for $r=0,\ldots,t-1$, leave
all other calibration rows fixed, and assign the free row to $k$.
The path is simple, so every row still has one column. Its initial column
$k$ becomes available, its terminal column $k^\star$ becomes occupied,
and every intermediate column remains occupied once. The resulting
assignment $M_P$ belongs to $\Sigma_k$ and satisfies
\begin{align*}
\ell_k-\ell_{k^\star}
&\leq\langle\widehat C,M_P\rangle
     -\langle\widehat C,M^\star\rangle\\
&=\sum_{r=0}^{t-1}
  \bigl(\widehat C_{i(j_r),j_{r+1}}-\widehat C_{i(j_r),j_r}\bigr)\\
&=\operatorname{wt}(P)=\Delta_k.
\end{align*}
For $k=k^\star$, take the empty path and $M_P=M^\star$.
Combining the bounds proves
$\ell_k-\ell_{k^\star}=\Delta_k$ and shows that $M_P$ attains $\ell_k$.
\end{proof}

\subsection{Dual variables and reduced displacement weights}

We now introduce the dual variables of the augmented assignment problem. The
linear programming relaxation of the augmented assignment problem is
\[
\min_{M\ge 0}\ \langle \widehat C,M\rangle
\]
subject to
\[
M\mathbf 1=\mathbf 1,
\qquad
M^\top\mathbf 1=\mathbf 1.
\tag{P}
\]
By the Birkhoff--von Neumann theorem, the feasible polytope is the convex hull
of the permutation matrices. Hence this relaxation has the same optimal value
as the augmented assignment problem.
The dual of (P) is \citep[\S2.5, Proposition~2.4]{peyre2019computational}
\[
\max_{u,v}\ \sum_{i=1}^{n+1} u_i+\sum_{j=1}^{n+1} v_j
\]
subject to
\[
u_i+v_j\le \widehat C_{ij},
\qquad i,j\in[n+1],
\tag{D}
\]
where $u,v\in\mathbb R^{n+1}$.
Let $(u^\star,v^\star)$ be dual-optimal variables, by definition
\[
u_i^\star+v_j^\star\le \widehat C_{ij}.
\]
Define the reduced costs
\[
\widetilde C_{ij}:=\widehat C_{ij}-u_i^\star-v_j^\star.
\]
By dual feasibility,
\[
\widetilde C_{ij}\ge 0
\qquad
\forall i,j\in[n+1].
\]

Because $M^\star$ is primal optimal and $(u^\star,v^\star)$ is dual
optimal, complementary slackness holds. Therefore, for every matched edge of
$M^\star$,
\[
M^\star_{ij}=1
\quad\Longrightarrow\quad
u_i^\star+v_j^\star=\widehat C_{ij}.
\]
In particular, for every $a\in \mathcal A^\star$,
\[
u_{i(a)}^\star+v_a^\star
=
\widehat C_{i(a),a}.
\]

Since the free row has zero costs, dual feasibility also includes
\[
u_{n+1}^\star+v_j^\star\le 0
\qquad
\forall j\in[n+1].
\]
Because the free row is matched to $k^\star$, complementary slackness gives
\[
u_{n+1}^\star+v_{k^\star}^\star=0.
\]
These free-row constraints are part of the full augmented dual system,
although the nonnegativity of the reduced displacement weights below uses only
the matched edges of the real rows.\looseness=-1

\paragraph{Dual non-uniqueness.}
The dual variables are not unique. This causes no ambiguity. The arguments
below hold for any dual-optimal pair $(u^\star,v^\star)$, because they use
only dual feasibility together with complementary slackness against the fixed
primal optimum $M^\star$. In particular, the standard gauge transformation
\[
u^\star\mapsto u^\star-\gamma\mathbf 1,
\qquad
v^\star\mapsto v^\star+\gamma\mathbf 1,
\qquad \gamma\in\mathbb R,
\]
leaves all reduced costs $\widetilde C_{ij}$, all reduced displacement weights
$\widetilde w$ defined below, and all differences
$v_{k^\star}^\star-v_k^\star$ unchanged. The independence of the recovery
formula from the choice of dual optimum is recorded at the end of the next
subsection.

Define the reduced displacement weight of the edge $a\to b$ by
\[
\widetilde w(a\to b)
:=
w(a\to b)+v_a^\star-v_b^\star.
\]

\begin{lemma}[Reduced displacement weights]
\label[lemma]{lem:reduced-displacement-weights}
For every displacement edge $a\to b$, with $a\in\mathcal A^\star$ and
$b\in[n+1]$,
$
\widetilde w(a\to b)
=
\widetilde C_{i(a),b}.
$
In particular,
$
\widetilde w(a\to b)\ge 0.
$
\end{lemma}

\begin{proof}
Complementary slackness on the matched edge $(i(a),a)$ gives
$\widehat C_{i(a),a}=u_{i(a)}^\star+v_a^\star$. Hence
\begin{align*}
\widetilde w(a\to b)
&=w(a\to b)+v_a^\star-v_b^\star\\
&=\widehat C_{i(a),b}-\widehat C_{i(a),a}
  +v_a^\star-v_b^\star\\
&=\widehat C_{i(a),b}-u_{i(a)}^\star-v_b^\star\\
&=\widetilde C_{i(a),b}\geq0,
\end{align*}
where the last inequality is dual feasibility.
\end{proof}

\subsection{Reduced-Cost Shortest Paths and Recovery}

Let $\widetilde\Delta_k$ denote the shortest-walk distance from $k$ to
$k^\star$ in the displacement graph with reduced edge weights
$\widetilde w$. Since
$
\widetilde w(a\to b)\ge 0
$
for every edge by \Cref{lem:reduced-displacement-weights}, every shortest
walk can be chosen simple: deleting a repeated cycle cannot increase the total
reduced weight. Therefore the shortest-walk and shortest-simple-path distances
coincide, and $\widetilde\Delta_k$ can be computed by Dijkstra's
algorithm.
For any directed walk
\[
W:\quad k=j_0\to j_1\to\cdots\to j_t=k^\star,
\]
we have
\[
\operatorname{wt}_{\rm red}(W)
=
\sum_{\ell=0}^{t-1}
\widetilde w(j_\ell\to j_{\ell+1}).
\]
Using the definition of $\widetilde w$,
\[
\begin{aligned}
\operatorname{wt}_{\rm red}(W)
&=
\sum_{\ell=0}^{t-1}
\left[
w(j_\ell\to j_{\ell+1})
+
v_{j_\ell}^\star
-
v_{j_{\ell+1}}^\star
\right] 
=
\operatorname{wt}(W)
+
v_k^\star-v_{k^\star}^\star.
\end{aligned}
\]
The potential terms telescope and depend only on the endpoints of the walk.
Taking minima over directed walks from $k$ to $k^\star$ gives
\[
\widetilde\Delta_k
=
\Delta_k
+
v_k^\star-v_{k^\star}^\star.
\]
Equivalently,
$
\Delta_k
=
\widetilde\Delta_k
+
v_{k^\star}^\star-v_k^\star.
$
Combining this identity with \Cref{thm:shortest-path-formula} gives the
recovery formula
\[
{
\ell_k
=
\ell_{k^\star}
+
\widetilde\Delta_k
+
v_{k^\star}^\star-v_k^\star.
}
\]

The quantity
$
\widetilde\Delta_k+v_{k^\star}^\star-v_k^\star
$
equals the raw distance $\Delta_k$. Hence the recovery formula is
independent of the particular dual optimum, although the reduced weights and
the reduced distances themselves may depend on that choice.